\pdfoutput=1

\documentclass[letterpaper, 10 pt, conference]{ieeeconf}  

\IEEEoverridecommandlockouts                              

\usepackage{graphics}
\usepackage[pdftex]{graphicx}
\usepackage[tight,footnotesize]{subfigure}
\usepackage{multirow}
\usepackage[cmex10]{amsmath}
\usepackage{amssymb}
\usepackage{amsfonts}
\usepackage{algpseudocode}
\usepackage{algorithm}
\usepackage{booktabs}
\usepackage{arydshln}
\usepackage{textcomp}
\usepackage{url}
\usepackage{color}
\usepackage[table]{xcolor}

\definecolor{firstcolor}{RGB}{120, 220, 115}
\colorlet{secondcolor}{green!25}
\colorlet{thirdcolor}{yellow!35}

\usepackage{siunitx}
\usepackage{soul}
\usepackage{xcolor}
\usepackage{colortbl}
\usepackage{wrapfig}
\usepackage{babel}
\usepackage[small]{caption}
\usepackage{capt-of}

\makeatletter
\newcommand{\cdashmidrule}[1]{%
  \noalign{\vskip\aboverulesep}
  \cdashline{#1}
  \noalign{\vskip\belowrulesep}}
\makeatother

\usepackage{array}
\usepackage{tcolorbox}
\usepackage[bookmarks=true,colorlinks]{hyperref}
\hypersetup{
  pdftitle={DGSG-Mind: Dynamic 3D Gaussian Scene Graphs for Long-Term Scene Understanding and Grounding},
  pdfauthor={Luzhou Ge, Xiangyu Zhu, Jinyan Liu, Xuesong Li}
}
\title{\LARGE \bf
DGSG-Mind: Dynamic 3D Gaussian Scene Graphs for Long-Term  \\ Scene Understanding and Grounding
}

\author{%
Luzhou Ge$^{1}$,
Xiangyu Zhu$^{1}$,
Jinyan Liu$^{1\ast}$, and Xuesong Li$^{1\ast}$%
\thanks{$^{1}$School of Computer Science, Beijing Institute of Technology, Beijing 100081, China.}%
\thanks{$^{\ast}$Corresponding authors: Xuesong Li ({\tt\footnotesize lixuesong@bit.edu.cn}) and Jinyan Liu ({\tt\footnotesize jyliu@bit.edu.cn}).}%
}

\makeatletter
\let\ArxivSupplementMaketitle\maketitle
\let\@oldmaketitle\@maketitle
\renewcommand{\@maketitle}{%
  \@oldmaketitle
  \begin{center}
    \includegraphics[width=\textwidth]{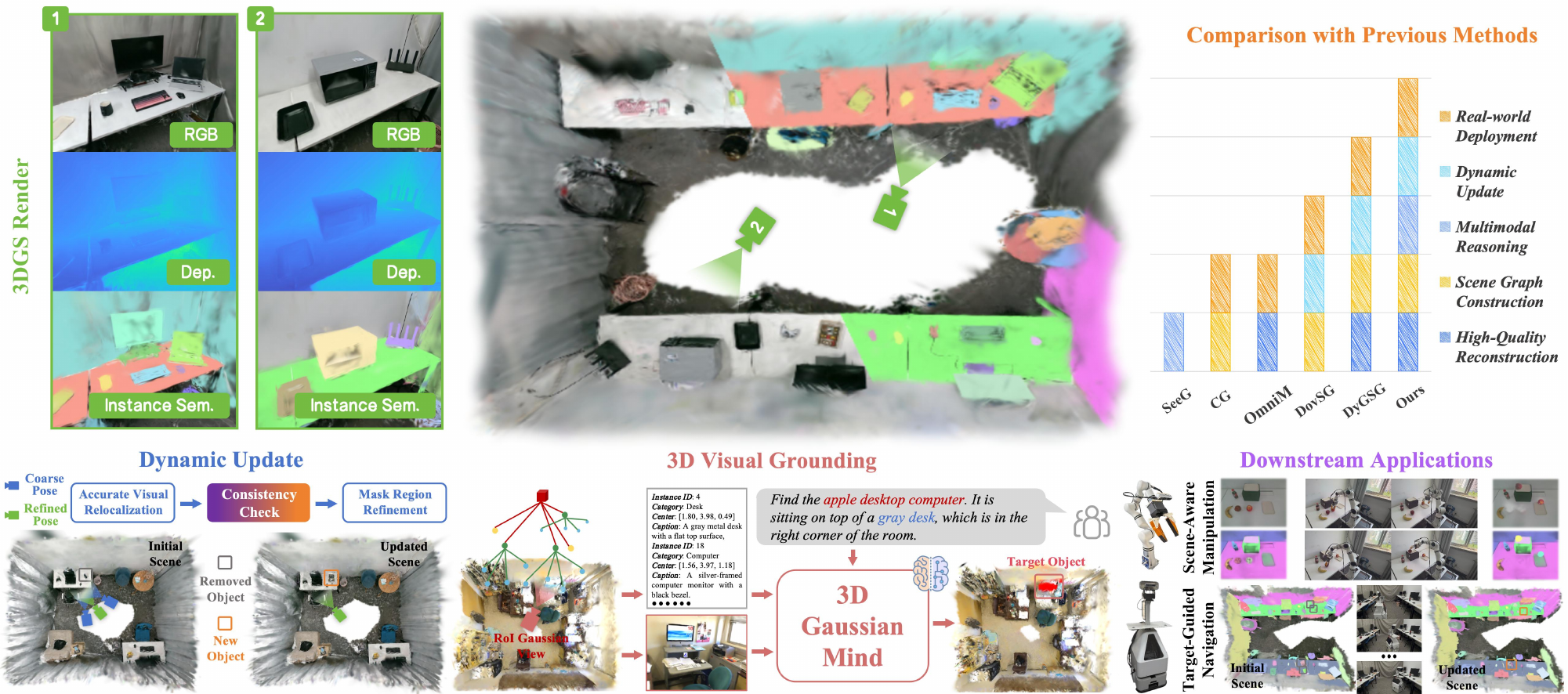}
    \captionof{figure}{Compared with previous state-of-the-art systems, \textbf{\textit{DGSG-Mind}} offers a more complete functional framework for long-term embodied scene understanding in dynamic environments. It jointly maintains a hybrid instance-aware 3D Gaussian representation and a hierarchical scene graph, supporting high-quality reconstruction, semantic mapping, accurate visual relocalization, and instance-level dynamic updates. Moreover, it leverages Region of Interest (RoI) Gaussian-rendered visual cues together with semantic and spatial relations for multimodal reasoning. These capabilities allow \textbf{\textit{DGSG-Mind}} to support long-term robotic tasks in dynamic real-world environments. \textit{Abbreviations:} SeeG (SeeGround~\cite{li2025seeground}), CG (ConceptGraphs~\cite{gu2023conceptgraphsopenvocabulary3dscene}), OmniM (OmniMap~\cite{deng2025omnimap}), DovSG~\cite{yan2025dynamicopenvocabulary3dscene}, and DyGSG (DynamicGSG~\cite{ge2025dynamicgsg}).}%
    \label{fig:teaser}%
  \end{center}
  \setcounter{figure}{1}%
  \vspace{-1em}
}
\makeatother

\begin{document}

\maketitle
 \thispagestyle{empty}
\pagestyle{empty}


\begin{abstract}

Integrating open-vocabulary semantic information into dynamic 3D scene representations is essential for long-term embodied scene understanding. However, existing methods often suffer from fragile instance association due to incomplete cross-view cues, while their limited ability to handle object-level topological changes restricts long-term robotic task execution. Moreover, current 3D scene understanding methods either rely on simple feature matching without explicit spatial reasoning or assume offline ground-truth 3D geometry. To address these challenges, we present \textbf{\textit{DGSG-Mind}}, a hybrid instance-aware 3D Gaussian dynamic scene graph system with an embodied reasoning agent. Our system couples a probabilistic voxel grid with explicit 3D Gaussians to enable robust cross-modal instance fusion and incremental semantic mapping. It handles dynamic changes through Gaussian-based visual relocalization and localized masked refinement guided by geometric, appearance, and semantic consistency. Built on the instance Gaussian map, DGSG-Mind further constructs a hierarchical scene graph and develops the \textbf{\textit{3D Gaussian Mind}}, which integrates structural relations, spatial-semantic information, and visually annotated RoI Gaussian renderings for multimodal reasoning. Extensive experiments show that DGSG-Mind achieves the best zero-shot 3DVG performance among methods operating on self-reconstructed maps, while also delivering strong performance in 3D open-vocabulary semantic segmentation and scene reconstruction. We further deploy \textbf{\textit{DGSG-Mind}} on real-world robots to demonstrate its target-oriented reasoning and dynamic update capabilities. The project page of \textbf{\textit{DGSG-Mind}} is available at \href{https://icr-lab.github.io/DGSG-Mind}{https://icr-lab.github.io/DGSG-Mind}.
\end{abstract}



\section{Introduction}
\label{sec:introduction}

Dynamic 3D semantic mapping and scene understanding are fundamental to embodied AI systems that execute long-term tasks in changing environments. Recently, 3D Gaussian Splatting (3DGS)~\cite{kerbl20233d} has been widely adopted for real-time dense mapping, while Vision-Language Models (VLMs) have enabled visual understanding and language-guided reasoning for embodied decision-making. However, integrating open-vocabulary semantics and spatial reasoning into dynamic scene representations remains challenging. Existing instance-level mapping methods rely on limited cross-view cues, such as 3D point-cloud overlap~\cite{gu2023conceptgraphsopenvocabulary3dscene, werby23hovsg} or simple 2D mask intersections~\cite{ge2025dynamicgsg, yang2025opengs-slam}, making them vulnerable to occlusions, detector instability, and multi-view inconsistencies. Moreover, most 3DGS-based semantic mapping frameworks~\cite{deng2025omnimap, yang2025opengs-slam, qin2024langsplat3dlanguagegaussian, yang2025opengs} assume static scenes, limiting their ability to handle object-level topological changes. Existing 3D scene understanding methods further rely either on simple feature similarity~\cite{qin2024langsplat3dlanguagegaussian, ye2024gaussiangroupingsegmentedit, zhu2025objectgs} without explicit spatial-topological reasoning, or offline ground-truth 3D point clouds for stronger spatial reasoning~\cite{li2025seeground, yuan2024visual,jin2025spazer}.



To address these limitations, we propose \textbf{\textit{DGSG-Mind}}, a dynamic 3D Gaussian scene graph framework for long-term scene understanding and grounding. It combines explicit 3D Gaussians with a sparse probabilistic voxel grid for incremental instance-aware mapping. The voxel grid accumulates voxel-level instance evidence and guides cross-view instance association and Gaussian densification, while the multi-term optimized Gaussian map provides renderable geometry, appearance, and instance semantics. Built on this map, \textbf{\textit{DGSG-Mind}} constructs a hierarchical scene graph and develops the \textit{3D Gaussian Mind}, which integrates relation-aware RoI renderings with structured scene information for zero-shot 3D visual grounding. For dynamic environments, Gaussian-based relocalization and joint consistency checks guide localized Gaussian updates and corresponding scene-graph synchronization.

In summary, the main contributions are as follows:

\begin{itemize}
    \item \textbf{Hybrid Instance-Aware 3DGS Representation:} We develop a hybrid voxel-Gaussian representation that integrates voxel-level instance evidence, rendered mask IoU, and semantic similarity for cross-view association, and supports voxel-guided densification and multi-term Gaussian optimization.
    
    \item \textbf{Geometric-Appearance-Semantic Dynamic Update:} We introduce a localized masked refinement strategy constrained by joint consistency to improve the accuracy of dynamic scene updates without re-optimizing the static background.
    
    \item \textbf{3D Gaussian Mind:} We construct a hierarchical scene graph and develop the \textit{3D Gaussian Mind} for multimodal reasoning. This module jointly leverages scene structure, node-level spatial-semantic information, and annotated RoI Gaussian renderings to support structured scene understanding and target reasoning in complex 3D scenes.
\end{itemize}

\section{Related Work}  \label{sec:related_work}

\subsection{Gaussian Dense Mapping}
\label{subsec:related_gaussian_mapping}

The development of 3D Gaussian Splatting (3DGS)~\cite{kerbl20233d} has significantly advanced real-time dense RGB-D SLAM (e.g., SplaTAM \cite{keetha2024splatam}, GS-ICP-SLAM \cite{ha2024rgbdicp}). To mitigate the geometric artifacts inherent in unconstrained 3DGS, hybrid representations such as GSFusion \cite{wei2024gsfusion} and SEGS-SLAM \cite{wen2025segs} successfully incorporate voxel grids to enforce structural regularizations. Recently, dense mapping works have been extended to embedding open-vocabulary semantics. By lifting semantic features from foundation models (e.g., SBERT \cite{reimers2019sentence}, CLIP \cite{radford2021learningtransferablevisualmodels}), methods like OmniMap \cite{deng2025omnimap}, DynamicGSG \cite{ge2025dynamicgsg}, and OpenGS-Fusion \cite{yang2025opengs} incrementally integrate semantic embeddings into a global map for open-vocabulary semantic mapping and achieve photorealistic reconstruction. However, their instance association strategies often remain fragile due to incomplete cross-view merging heuristics. By contrast, our method integrates a probabilistic voxel grid with Gaussian-rendered 2D instance masks for cross-modal instance association, improving robustness and accuracy to object occlusions and instance fusion.

\subsection{3D Dynamic Scene Graphs}
\label{subsec:related_scene_graph}

3D scene graphs~\cite{hughes2024foundations} represent spatial geometry and semantic relationships as hierarchical topological graphs. With the development of vision-language models, methods such as ConceptGraphs~\cite{gu2023conceptgraphsopenvocabulary3dscene}, HOVSG~\cite{werby23hovsg}, and BBQ~\cite{linok2024barequeriesopenvocabularyobject} lift open-vocabulary semantics into 3D object nodes for downstream robotic tasks. To handle environmental changes, RoboExp~\cite{jiang2024roboexpactionconditionedscenegraph}, DovSG~\cite{yan2025dynamicopenvocabulary3dscene}, and DynamicGSG~\cite{ge2025dynamicgsg} update scene graphs by modeling robot actions or detecting scene changes. However, RoboExp~\cite{jiang2024roboexpactionconditionedscenegraph} mainly focuses on action-conditioned interactive exploration rather than long-term scene reconstruction. DovSG~\cite{yan2025dynamicopenvocabulary3dscene} relies on a multi-stage relocalization pipeline with Accelerated Coordinate Encoding (ACE)~\cite{brachmann2023accelerated}, feature matching, and point-cloud ICP, making its accuracy dependent on geometric registration. DynamicGSG~\cite{ge2025dynamicgsg} requires external VIO poses and still optimizes unaffected regions during scene updates. Moreover, these systems rely on geometric or photometric consistency for change detection, which can be insufficient under appearance changes, occlusion, or ambiguous geometry. In contrast, our method performs dynamic updates with Gaussian-based relocalization, joint geometric, appearance, and semantic consistency, and localized masked refinement, enabling accurate scene updates without re-optimizing the static background.

\subsection{3D Scene Understanding}
\label{subsec:scene_understanding}

3D Visual Grounding (3DVG), which localizes target objects from free-form language descriptions, is an important benchmark for embodied scene understanding. Recent methods increasingly use LLMs and MLLMs as reasoning agents for this task. Methods such as LLM-Grounder~\cite{yang2024llm} and ZSVG3D~\cite{yuan2024visual} translate language queries into spatial heuristics for object search, while more advanced methods (e.g., SeeGround~\cite{li2025seeground}, SPAZER~\cite{jin2025spazer})  further incorporate structural guidance and spatial-semantic reasoning. However, existing methods still face two limitations. Open-vocabulary 3D representations such as  LangSplat~\cite{qin2024langsplat3dlanguagegaussian}, GaussianGrouping~\cite{ye2024gaussiangroupingsegmentedit}, and ObjectGS~\cite{zhu2025objectgs} mainly rely on simple feature matching and lack explicit spatial-topological structure. In contrast, spatial reasoning agents~\cite{li2025seeground,yuan2024visual,jin2025spazer} are designed for offline settings with pre-scanned geometry. Our method performs zero-shot 3DVG directly on self-reconstructed Gaussian maps, using rendered RoI views and structured scene-graph contexts for visual-spatial reasoning.



\begin{figure*}
  \centering
  \includegraphics[width=1.0\textwidth]{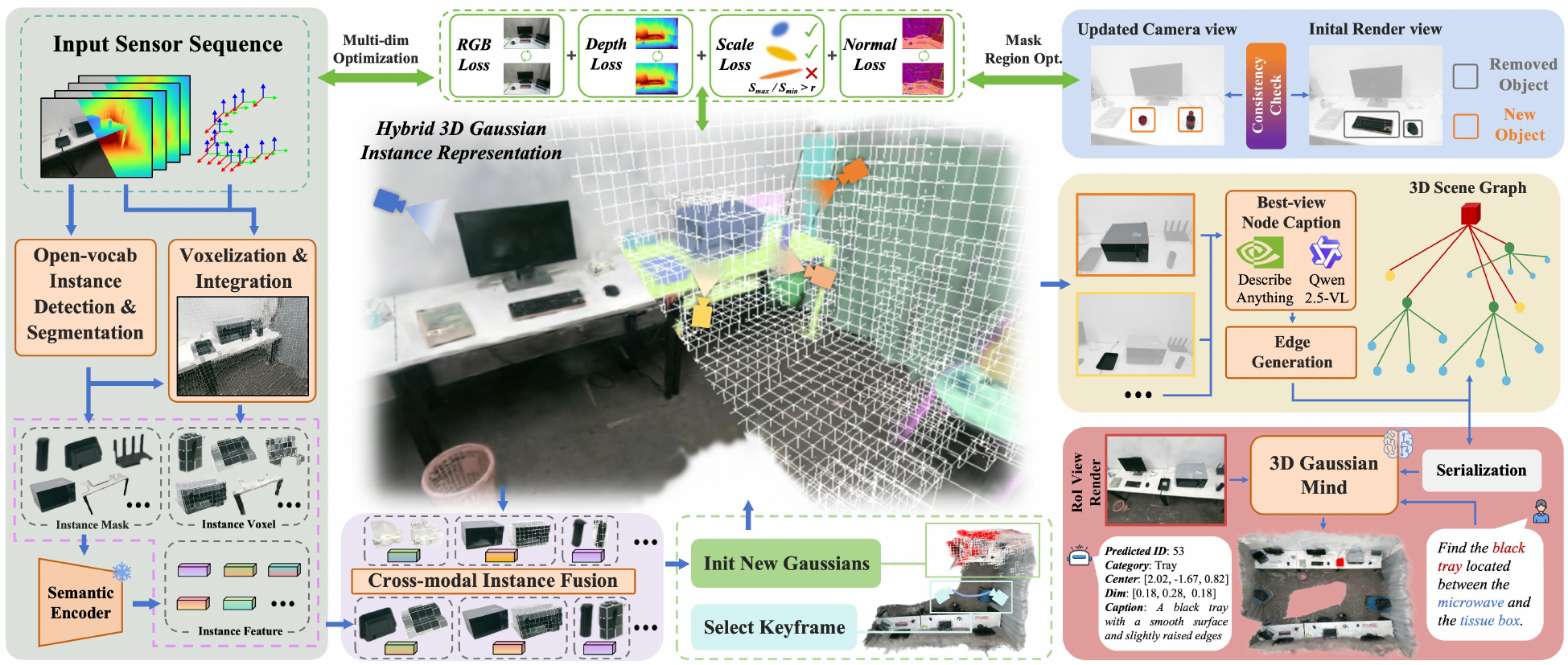}
    \caption{\textbf{\textit{System Overview of DGSG-Mind.}} Given a posed RGB-D sequence, \textit{DGSG-Mind} extracts open-vocabulary instance masks and semantic features, and integrates them into a hybrid 3D Gaussian instance representation. Cross-modal association leverages a sparse probabilistic voxel grid, rendered Gaussian masks, and instance-level features to link 2D observations with persistent 3D Gaussian instances. The sparse probabilistic voxel grid also guides Gaussian initialization, while the Gaussian field is optimized across multiple views with photometric, depth, scale, and normal regularization. The instance-aware Gaussian map is further abstracted into a hierarchical 3D scene graph, enabling the \textit{3D Gaussian Mind} to perform zero-shot 3D visual grounding and spatial reasoning from rendered RoI views and structured 3D relations. For dynamic scenarios, localized masked refinement updates newly appeared or removed objects while keeping the unchanged background fixed.
}
  \label{fig:pipeline}
  \vspace{-1.5em}
\end{figure*}

\section{Methodology}
\label{sec:methodology}
An overview of \textbf{\textit{DGSG-Mind}} is shown in Fig.~\ref{fig:pipeline}. 
Our system builds a hybrid instance-aware 3D Gaussian representation by coupling a sparse probabilistic voxel grid with explicit Gaussians. 


\subsection{3D Gaussian Preliminaries}
\label{para:gaussian_preliminary}

A scene is represented as a set of anisotropic 3D Gaussians,
\(
\mathcal{G}=\{G_i\}_{i=1}^{N}
\).
Each Gaussian \(G_i\) is parameterized by its center
\(\mathbf{u}_i \in \mathbb{R}^3\),
color attribute
\(\mathbf{c}_i \in \mathbb{R}^3\),
scale
\(\mathbf{s}_i \in \mathbb{R}^3\),
rotation
\(\mathbf{R}_i \in SO(3)\),
covariance
\(\mathbf{\Sigma}_i \in \mathbb{R}^{3\times 3}\),
and opacity
\(o_i \in [0,1]\).
The covariance is constructed from the scale and rotation as
\(
\mathbf{\Sigma}_i=\mathbf{R}_i \operatorname{diag}(\mathbf{s}_i^2)\mathbf{R}_i^\top.
\) The unnormalized spatial density of \(G_i\) is defined as: 
\begin{equation}
\mathcal{N}_i(\mathbf{x})
=
\exp\!\left(
-\frac{1}{2}
(\mathbf{x}-\mathbf{u}_i)^\top
\mathbf{\Sigma}_i^{-1}
(\mathbf{x}-\mathbf{u}_i)
\right).
\end{equation} 

Given a camera pose \(\mathbf{T} \in SE(3)\), each Gaussian is projected onto the image plane with projected covariance
\(\mathbf{\Sigma}'_i=\mathbf{J}_i \mathbf{T}^{-1}\mathbf{\Sigma}_i \mathbf{T}^{-T}\mathbf{J}_i^\top\),
where \(\mathbf{J}_i\) denotes the Jacobian of the projection function. The rendered color and depth at pixel \(p\) are computed by alpha blending: \(
C(p)=\sum_{i=1}^{N} T_i(p)\,\alpha_i(p)\,\mathbf{c}_i, \) \(
D(p)=\sum_{i=1}^{N} T_i(p)\,\alpha_i(p)\,d_i,
\) where \(\alpha_i(p)\) denotes the alpha contribution of the \(i\)-th Gaussian at pixel \(p\), \(T_i(p)=\prod_{j=1}^{i-1}(1-\alpha_j(p))\) denotes the accumulated transmittance along the ray, and \(d_i\) represents the depth of the Gaussian center in the camera coordinate.

\subsection{Voxel-based Instance-Aware 3DGS Mapping}
\label{instance_aware_3DGS}

Given a posed RGB-D stream \(\mathcal{I}_t=\langle C_t, D_t, \mathbf{T}_t \rangle \), our system incrementally constructs a hybrid instance-aware map consisting of explicit 3D Gaussians and a sparse probabilistic voxel grid. For each frame \(\mathcal{I}_t\), we first identify the active voxel blocks within the current view frustum and integrate the RGB-D observation into the sparse voxel map. Valid depth pixels are then back-projected into 3D and associated with their corresponding voxel blocks and local voxel indices. Each voxel maintains candidate instance IDs and accumulated hit statistics, from which voxel-wise assignment probabilities are derived. This hybrid representation serves as an efficient geometric scaffold for incrementally linking 2D observations with persistent 3D Gaussian instances.

\paragraph{Instance Recognition}
For each observation \(\mathcal{I}_t\), we employ the open-vocabulary object detector YOLO-World~\cite{cheng2024yoloworldrealtimeopenvocabularyobject} to extract detection bounding boxes \(\{b_{t,i}\}_{i=1}^{M}\). To obtain precise instance boundaries, these detections are fed into the Segment Anything Model ~\cite{kirillov2023segment}, which produces high-quality 2D instance masks \(\{m_{t,i}\}_{i=1}^{M}\). For each mask, CLIP~\cite{radford2021learningtransferablevisualmodels} is used to extract a normalized semantic feature \(\mathbf{f}_i^t\). Accordingly, the 2D observation of the \(i\)-th instance at timestep \(t\) is represented as \(O_i^t = (m_i^t, \mathbf{f}_i^t)\).

\paragraph{Cross-modal Instance Association}

For each 2D instance observation \(O_i^t\), we back-project the mask \(m_i^t\) into 3D using \(D_t\) and map the corresponding points into the voxel grid with camera pose \(\mathbf{T}_t\). Let \(\mathcal{V}_i^t\) denote the set of valid voxels occupied by the observation \(O_i^t\). Each voxel stores up to three candidate instance IDs with accumulated hit counts. For voxel \(v\), we define the assignment probability of candidate instance \(I_k\) as \(P_k(v)=\frac{c_k(v)}{c(v)}\) when \(I_k\) is stored in \(v\), and \(P_k(v)=0\) otherwise, where \(c_k(v)\) and \(c(v)\) denote the hit count of \(I_k\) and the total hit count in \(v\), respectively. Querying the stored historical instance IDs within \(\mathcal{V}_i^t\) yields the candidate map instance set \(\mathcal{C}_i^t=\{I_k\}\). We compute three similarity terms to associate \(O_i^t\) with each candidate \(I_k\in\mathcal{C}_i^t\):

\begin{itemize}
    \item \textbf{Geometric Similarity (\(S_{\mathrm{geo}}(i,k)\)):}
    We measure the probabilistic spatial overlap in the voxel grid as \(S_{\mathrm{geo}}(i,k)=\frac{1}{|\mathcal{V}_i^t|}\sum_{v\in\mathcal{V}_i^t} P_k(v),\) where \(|\mathcal{V}_i^t|\) denotes the number of voxels occupied by \(O_i^t\) at time \(t\).
    \item \textbf{Rendered Mask IoU (\(S_{\mathrm{iou}}(i,k)\)):}
    We render the 3D Gaussians associated with \(I_k\) into the current view \(t\) to obtain a projected mask \(m_{k\rightarrow t}\), and compute \(S_{\mathrm{iou}}(i,k)=\frac{|m_{k\rightarrow t}\cap m_i^t|}{|m_{k\rightarrow t}\cup m_i^t|}\).
    \item \textbf{Semantic Similarity (\(S_{\mathrm{sem}}(i,k)\)):}
    We compute the cosine similarity between the current semantic feature \(\mathbf{f}_i^t\) and the maintained global feature \(\mathbf{F}_k\) of instance \(I_k\): \(S_{\mathrm{sem}}(i,k)=\frac{\mathbf{f}_i^t\cdot\mathbf{F}_k}{\|\mathbf{f}_i^t\|\,\|\mathbf{F}_k\|}\).
\end{itemize}

These three terms are combined into a joint similarity:
\begin{equation}
S(i,k)=\lambda_{\mathrm{1}}S_{\mathrm{geo}}(i,k)+\lambda_{\mathrm{2}}S_{\mathrm{iou}}(i,k)+\lambda_{\mathrm{3}}S_{\mathrm{sem}}(i,k),
\end{equation}
where \(\lambda_{\mathrm{1}} = 0.4\), \(\lambda_{\mathrm{2}} = 0.4\), and \(\lambda_{\mathrm{3}} = 0.2\). The observation \(O_i^t\) is associated with the instance \(k^*=\arg\max_k S(i,k)\) if the maximum score exceeds a threshold \(\tau\). Otherwise, \(O_i^t\) is initialized as a new instance.

\paragraph{Instance Fusion}
After associating \(O_i^t\) with a map instance \(I_{k^*}\), we update its semantic feature by: 
\begin{equation}
\mathbf{F}_{k^*} = 
\frac{\mathbf{F}_{k^*}W_{k^*}+\mathbf{f}_i^t w_i^t}{W_{k^*}+w_i^t},
\end{equation}
where \(w_i^t=(|\mathcal{V}_i^t|/{\hat{V}_{k^*})} \cdot S(i,k^*)\) is a visibility weight, \(\hat{V}_{k^*}\) denotes the total number of valid voxels currently assigned to \(I_{k^*}\) in the global voxel map, and \(W_{k^*}\) is the accumulated feature weight. We then update the voxel statistics: if \(k^*\) is already stored in voxel \(v\), its hit count \(c_{k^*}(v)\) is incremented; otherwise, \(k^*\) is inserted into an empty slot when available and updated accordingly. The total voxel count \(c(v)\) is incremented for all \(v\in\mathcal{V}_i^t\).

\subsection{Multi-dimensional Gaussian Optimization}
\label{subsec:gaussian_optim}

\paragraph{Voxel-based Gaussian Densification}
\label{para:densification}
Our method incrementally densifies the map from the probabilistic voxel grid. For each observation \(\mathcal{I}_t\), we examine valid voxels within the current camera frustum and instantiate new Gaussians only for newly observed voxels. Each voxel center \(\mathbf{x}\in\mathbb{R}^3\) initializes a Gaussian with center \(\mathbf{u}=\mathbf{x}\), identity rotation, and initial log-scale \(s_0\). The color is initialized from the zero-order SH coefficient of the observed RGB value, and the opacity is set to \(0.5\) via inverse-sigmoid mapping. We further maintain an explicit voxel-to-Gaussian index table to efficiently retrieve voxel-associated Gaussian subsets for local rendering and accelerated mask-IoU computation during instance association.

\paragraph{Scale \& Normal Regularization}
\label{para:regularization}
In unconstrained 3DGS optimization, Gaussians may become overly elongated or excessively large in textureless regions, causing geometric distortion and floating artifacts. To alleviate this issue, we employ two complementary regularizers on Gaussian scale and local surface geometry. Specifically, we introduce a scale anisotropy regularization that constrains the ratio between the maximum and minimum scales of each Gaussian:
\begin{equation}
\mathcal{L}_{scale} = \frac{1}{|\mathcal{G}^*|} \sum_{G \in \mathcal{G}^*} \max \left(0, \log\left(\frac{\max(\mathbf{s})}{\min(\mathbf{s}) \cdot r_{allow}}\right) \right),
\end{equation}
where $r_{allow} = 10$ is the maximum allowed aspect ratio, and $\mathcal{G}^*$ denotes the subset of violating Gaussians. We further apply a normal consistency regularization to stabilize local geometry by penalizing the angular discrepancy between the pseudo-ground-truth normals $\mathbf{N}_t$, computed from the spatial gradients of the input depth map, and the normals $\hat{\mathbf{N}}_t$ reconstructed from rendered depth at viewpoint $t$: 
\begin{equation}
\mathcal{L}_{normal} = 1 - \langle \hat{\mathbf{N}}_t, \mathbf{N}_t \rangle, 
\end{equation}
where $\langle \cdot, \cdot \rangle$ denotes the inner product between normalized normal vectors. 

\paragraph{Joint Optimization}
\label{para:map_optimization}
Following \cite{kerbl20233d} and \cite{keetha2024splatam}, we minimize a multi-term loss combining photometric and geometric regularization. The photometric term compares the rendered image $\hat{\mathbf{C}}_t$ with the RGB observation $\mathbf{C}_t$ at viewpoint $t$ using a combination of L1 loss and SSIM. For geometric supervision, we compute an L1 depth loss between the rendered depth map $\hat{\mathbf{D}}_t$ and the sensor depth map $\mathbf{D}_t$:
\begin{equation}
\mathcal{L}_{rgb} = (1 - \lambda_4) \| \hat{\mathbf{C}}_t - \mathbf{C}_t \|_1 + \lambda_4 (1 - \text{SSIM}(\hat{\mathbf{C}}_t, \mathbf{C}_t)),
\end{equation}
\begin{equation}
\mathcal{L}_{depth} = \| \hat{\mathbf{D}}_t - \mathbf{D}_t \|_1.
\end{equation}

The joint optimization loss is defined as:
\begin{equation}
\mathcal{L}_{total} = \mathcal{L}_{rgb} + \lambda_{5} \mathcal{L}_{depth} + \lambda_{6} \mathcal{L}_{normal} + \lambda_{7} \mathcal{L}_{scale},
\end{equation}
where $\lambda_{4} = 0.2$, $\lambda_{5} = \lambda_{6} = 0.1$, and $ \lambda_{7} = 1.0$. During optimization, we prune redundant Gaussians with extreme transparency or excessive spatial extent, and perform multi-view scene optimization over a keyframe set. Keyframes are selected every \(\delta_n\) frames or whenever the camera translation or rotation exceeds a preset threshold, and \(\delta_m\) historical keyframes are chosen for joint multi-view optimization.

\subsection{3D Scene Graph Construction}
\label{subsec:scene_graph}

After optimizing the instance-aware 3D Gaussian map, we construct a hierarchical 3D scene graph \(\mathcal{S}=(\mathcal{O},\mathcal{E})\), where \(\mathcal{O}=\{o_j\}_{j=1}^{M}\) is the set of object nodes and \(\mathcal{E}\) is the set of edges. Each node \(o_j=(\mathbf{p}_j,\mathbf{b}_j,\ell_j,\kappa_j,\nu_j,\delta_j)\) is associated with a 3D center \(\mathbf{p}_j\), a 3D bbox \(\mathbf{b}_j\), an object category \(\ell_j\), a caption \(\kappa_j\), a best view \(\nu_j\), and a functional role \(\delta_j\in\{\textit{Asset}, \textit{Ordinary}, \textit{Standalone}\}\). For each instance \(j\), we compute the centroid of the camera centers of all historical views in which it is observed and retain the top-\(K=5\) nearest views, ranked by the Euclidean distance from each view's camera center to this centroid. The best view \(\nu_j\) is selected as the one with the highest view quality score \(q_{u,j}=|m_{u,j}|/|\Omega_u|\cdot |\mathcal{H}_{u,j}|/|\mathcal{H}_j|\), where \(|\Omega_u|\) is the number of pixels in view \(u\), \(|m_{u,j}|\) is the number of mask pixels of instance \(j\), \(\mathcal{H}_j\) is the set of Gaussians associated with instance \(j\), and \(\mathcal{H}_{u,j}\subseteq\mathcal{H}_j\) is the visible subset in view \(u\). Following \cite{ge2025dynamicgsg}, \textit{Asset} denotes floor-standing furniture that can support or store objects, \textit{Standalone} denotes independently rooted objects or fixtures, and \textit{Ordinary} covers the remaining, typically movable objects. Based on the masked image from the selected best view, Describe Anything \cite{lian2025describe} first generates a fine-grained physical caption \(\kappa_j\), and Qwen2.5-VL-72B-Instruct \cite{bai2025qwen25vltechnicalreport} further parses it into the node category \(\ell_j\) and functional role \(\delta_j\). We then organize all nodes into a hierarchical topology using a semantic-spatial constraint based on their functional roles, node centers, and 3D bboxes. Specifically, an \textit{Asset}-to-\textit{Ordinary} edge \((o_a,o_b)\in\mathcal{E}\) is formed when \(o_b\)'s center is horizontally within \(o_a\)'s bbox and vertically inside or above it, representing containment or support relations, respectively. \textit{Asset} and \textit{Standalone} nodes connect directly to the root; each \textit{Ordinary} node connects to a compatible \textit{Asset} parent, or directly to the root if no compatible parent is found.
\subsection{Dynamic Scene Update}
\label{subsec:dynamic_update}

\paragraph{Gaussian-based Camera Relocalization}
Accurate camera relocalization is essential for long-term scene graph maintenance and task execution. Specifically, we first train a scene-specific
ACE model~\cite{brachmann2023accelerated} using posed RGB images from the current scene. Given a new observation, ACE predicts a coarse pose \(\mathbf{T}_{\mathrm{coarse}}\in SE(3)\), which is further refined by freezing the Gaussian map and optimizing only the camera pose. To improve robustness, we use a silhouette mask to retain sufficiently visible pixels and minimize the masked RGB-D alignment loss:
\begin{equation}
\scalebox{0.9}{$\displaystyle
\mathcal{M}_{\mathrm{r}}(p)
=
\mathbb{I}\!\left(
\hat{\mathbf{D}}(p)>0,\,\mathbf{D}(p)>0,\,
\sum_i T_i(p)\alpha_i(p)>\lambda_8
\right),
$}
\end{equation}
\vspace{-0.5em}
\begin{equation}
\mathcal{L}_{\mathrm{reloc}}
=
\mathcal{M}_{\mathrm{r}}
\odot
\left(
\|\hat{\mathbf{C}}-\mathbf{C}\|_1
+
\lambda_{9}\|\hat{\mathbf{D}}-\mathbf{D}\|_1
\right),
\end{equation}
where \(\lambda_8=0.98\) is the silhouette threshold and \(\lambda_9=0.5\). The pose with the minimum relocalization loss is selected as the refined pose \(\mathbf{T}_{\mathrm{refined}}\).

\paragraph{Dynamic Detection and Refinement}

Unlike DovSG \cite{yan2025dynamicopenvocabulary3dscene} and DynamicGSG \cite{ge2025dynamicgsg}, which rely primarily on geometric similarity for change detection, our method jointly evaluates geometric, appearance, and semantic consistency under the refined camera pose \(\mathbf{T}_{\mathrm{refined}}\). We consider only instances with more than \(50\%\) visible associated Gaussians in the current frustum. For each visible instance \(I_j\), let \(m_{t,j}\) be its rendered mask, \(\Omega_{t,j}=\{p\mid p\in m_{t,j},\,\hat{\mathbf{D}}_t(p)>0,\,\mathbf{D}_t(p)>0\}\) be the valid depth region, and \(\hat{\mathbf{g}}_j,\mathbf{g}_j\) be the CLIP features extracted from the rendered and observed RGB crops. We compute geometric, appearance, and semantic consistency as \(S_{\mathrm{geo
}}(j)=\sum_{p\in\Omega_{t,j}}\mathbb{I}(|\hat{\mathbf{D}}_t(p)-\mathbf{D}_t(p)|<\tau_d) / |\Omega_{t,j}|\), \(S_{\mathrm{app}}(j)=\mathrm{SSIM}(\hat{\mathbf{C}}_t(m_{t,j}),\mathbf{C}_t(m_{t,j}))\), and \(S_{\mathrm{sem}}(j)=\hat{\mathbf{g}}_j^\top\mathbf{g}_j/(\|\hat{\mathbf{g}}_j\|\,\|\mathbf{g}_j\|)\). The final change similarity is:
\begin{equation}
S_{\mathrm{change}}(j)=\lambda_{10}S_{\mathrm{geo}}(j)+\lambda_{11}S_{\mathrm{app}}(j)+\lambda_{12}S_{\mathrm{sem}}(j),
\end{equation}
where \(\tau_d = 0.05\,\mathrm{m}, \lambda_{10} = 0.2, \lambda_{11}=0.4, \lambda_{12}=0.4\). If \(S_{\mathrm{change}}(j)<\delta_{\mathrm{change}}\), we regard \(I_j\) as inconsistent with the current observation and add it to the removal set \(\mathcal{O}_{\mathrm{remove}}\). Its associated Gaussian primitives are then pruned from the map. After removing inconsistent instances, we perform residual instance detection on the current RGB-D observation following Sec.~\ref{instance_aware_3DGS} to identify newly appeared objects. The newly detected regions are back-projected into 3D and initialized as new Gaussian instances \(\mathcal{O}_{\mathrm{new}}\). Let \(\mathcal{M}_{\mathrm{new}}\) and \(\mathcal{M}_{\mathrm{remove}}\) denote the union masks of newly added and removed objects. We refine the updated Gaussian map only within the local update region \(\mathcal{M}_{\mathrm{update}}=\mathcal{M}_{\mathrm{new}}\cup\mathcal{M}_{\mathrm{remove}}\), while keeping the static background fixed following Sec.~\ref{subsec:gaussian_optim}. This localized masked refinement enables effective adaptation to dynamic updates without re-optimizing the entire scene.

\begin{figure}[t]
  \centering
  \includegraphics[width=1.0\linewidth]{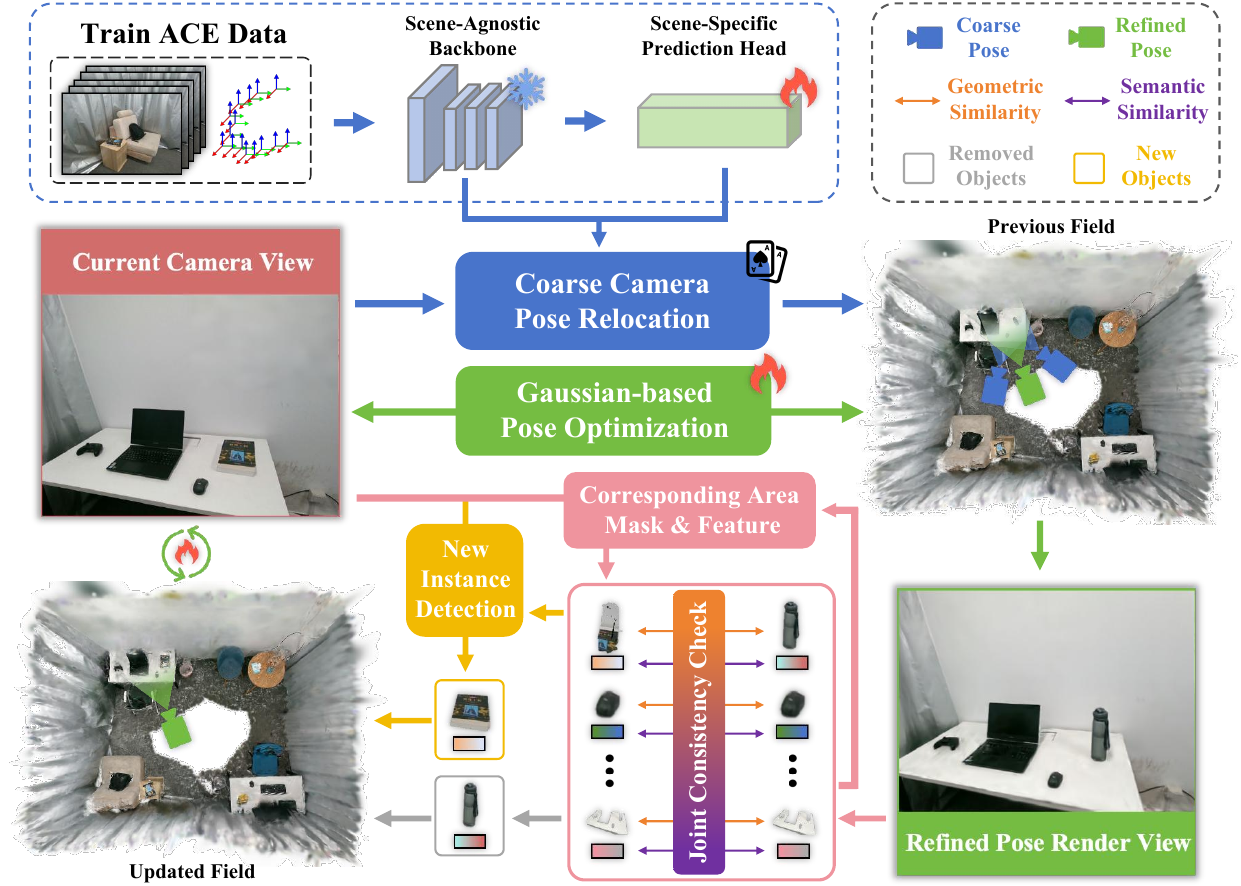}
  \caption{\textbf{Dynamic Scene Update:} Given the current camera view, we first estimate a coarse camera pose using a scene-specific ACE model and refine the pose against the 3D Gaussian map. With the refined pose, visible instances are evaluated using joint consistency to detect removed objects, while residual detection identifies newly appeared ones. The Gaussian map is then optimized by localized masked refinement, and the scene graph is synchronized with the resulting object additions and removals.}
  \label{fig:dynamic_update}
  \vspace{-1.5em}
\end{figure}

\paragraph{Scene Graph Update}
Let \(\mathcal{S}^{t-1}=(\mathcal{O}^{t-1},\mathcal{E}^{t-1})\) and \(\mathcal{S}^{t}=(\mathcal{O}^{t},\mathcal{E}^{t})\) denote the scene graph before and after the dynamic update. We update \(\mathcal{S}^{t-1}\) using the detected additions \(\mathcal{O}_{\mathrm{new}}\) and removals \(\mathcal{O}_{\mathrm{remove}}\). For each removed instance \(I_j\in\mathcal{O}_{\mathrm{remove}}\), the corresponding Standalone or Ordinary node \(o_j\) and its incident edges are deleted, while an Asset node is removed together with its associated subgraph. For each new instance \(I_j\in\mathcal{O}_{\mathrm{new}}\), we create a node \(o_j=(\mathbf{p}_j,\mathbf{b}_j,\ell_j,\kappa_j,\nu_j,\delta_j)\) and establish edges according to Sec.~\ref{subsec:scene_graph}. This keeps the Gaussian scene graph consistent with the updated scene state. Moved objects follow this remove-and-add path and receive a new identity.

\subsection{3D Gaussian Mind}
\label{subsec:gs_mind}

\begin{figure}[t]
  \centering
  \includegraphics[width=1.0\linewidth]{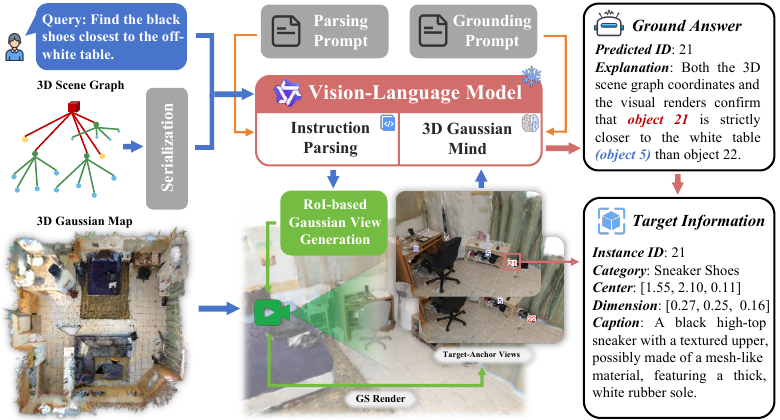}
  \caption{\textbf{3D Gaussian Mind:} By integrating natural language queries, structured 3D scene graphs, and generated annotated Gaussian views (RoI images), this framework leverages a Vision-Language Model for joint spatial reasoning and object localization.}
  \label{fig:GS_mind}
  \vspace{-1.5em}
\end{figure}

\paragraph{Instruction Parsing}
To support zero-shot 3D visual grounding and spatial reasoning from free-form language instructions, we introduce the \textit{3D Gaussian Mind}, illustrated in Fig.~\ref{fig:GS_mind}. Given a natural language query, we use a VLM-based parser to extract the target object (\textit{Target}) and its reference object(s) (\textit{Anchor}). The 3D scene graph is serialized in JSON format as structured context for subsequent multimodal grounding. Each parsed Target or Anchor phrase \(q\) is independently matched to an object node \(o_j\) using the cosine score \(s(q,o_j)\) between the phrase's CLIP feature and \(o_j\)'s associated map-instance feature. The corresponding candidate set contains up to five top-scoring nodes satisfying \(s(q,o_j)\geq\max(0.2,s_{\max}(q)-0.1)\), where \(s_{\max}(q)=\max_j s(q,o_j)\).


\paragraph{RoI-based Gaussian View Generation}
For each Target candidate \(o_j\), we group it with all retrieved Anchor candidates other than the same instance into the relation-aware region \(\mathcal{R}_j\). Starting from the Target's best-view \(\nu_j\), the virtual camera extrinsic is constructed as \(\nu_j^{\mathrm{RoI}}=\operatorname{LookAtFit}(\nu_j,\mathcal{R}_j)\). Here, \(\operatorname{LookAtFit}\) uses the best-view extrinsic only as a scene-consistent initialization and iteratively re-orients and translates the virtual camera, under fixed intrinsics, to fit the Target--Anchor 3D-bbox projections within the image as completely as possible while constraining the camera center to the mapped scene's XY bounds. We render the Gaussian map from the resulting pose and annotate the projected Target and Anchor instances with numerical IDs following~\cite{li2025seeground}.

\paragraph{Mind-guided Visual Grounding}
Finally, given the original query and the Target and Anchor candidate sets, the VLM selects a target object from the Target set by jointly reasoning over the ID-annotated Gaussian views and the hierarchically serialized JSON scene graph, combining visual appearance with semantic-spatial context.

\section{Experiments}
\label{sec:experiment}
\subsection{Experimental Setup}
\label{subsec:experiment_setup}

\textbf{Evaluations:} To comprehensively evaluate \textit{DGSG-Mind}, we conduct experiments across four key tasks: 1) \textit{3D open-vocabulary semantic segmentation}, which assesses instance-level, zero-shot 3D scene understanding; 2) \textit{3D visual grounding}, which evaluates the system's ability to localize target objects given free-form natural language descriptions; 3) \textit{scene reconstruction}, which measures the photometric fidelity of the optimized 3D Gaussian representation; 4) \textit{real-world deployment}, which validates its capability to support continuous robot operation, target-oriented navigation, and dynamic scene updates in real-world environments. 

\textbf{Datasets:} For open-vocabulary semantic segmentation and scene reconstruction, we use all eight synthetic scenes from Replica~\cite{replica19arxiv} and eight ScanNet~\cite{dai2017scannet} scenes selected following BBQ~\cite{linok2024barequeriesopenvocabularyobject}: \texttt{scene0011\_00}, \texttt{scene0030\_00}, \texttt{scene0046\_00}, \texttt{scene0086\_00}, \texttt{scene0222\_00}, \texttt{scene0378\_00}, \texttt{scene0389\_00}, and \texttt{scene0435\_00}. For 3DVG, we use ScanRefer~\cite{chen2020scanrefer} and Nr3D~\cite{achlioptas2020referit3d} queries from the same ScanNet scenes.  

\textbf{Implementation Details:} All experiments are conducted on a workstation with an NVIDIA RTX 4090D GPU, with \(\delta_n=5\), \(\delta_m=10\), and \(\delta_{\mathrm{change}}=0.35\). The sparse probabilistic voxel grid uses a resolution of \(0.03\,\mathrm{m}\). The translation and rotation thresholds for keyframe selection are set to \(0.1\,\mathrm{m}\) and \(0.1\,\mathrm{rad}\), respectively. For fair comparison, all agent-based methods requiring a Vision-Language Model (VLM) use \texttt{Qwen2.5-VL-72B-Instruct}~\cite{bai2025qwen25vltechnicalreport}.

\subsection{3D Open-Vocabulary Semantic Segmentation}
\label{subsec:semantic_segmentation}

\renewcommand{\arraystretch}{1.2}
\setlength{\tabcolsep}{1.7pt}
\begin{table}[!t]
    \centering
    \begin{tabular}{lcccccc}\toprule
\multirow{2.5}{*}{\textbf{Methods}}      & \multicolumn{3}{c}{\textbf{Replica}} & \multicolumn{3}{c}{\textbf{ScanNet}} \\
\cmidrule(lr){2-4} \cmidrule(lr){5-7} 
& mAcc$\uparrow$        & mIoU$\uparrow$   & F-mIoU$\uparrow$  & mAcc$\uparrow$        & mIoU$\uparrow$   & F-mIoU$\uparrow$  \\ \midrule 

ConceptGraphs \cite{gu2023conceptgraphsopenvocabulary3dscene}  & 40.78  & 26.28   & 42.49     & 41.31      & 22.02   & 30.90  \\
OpenGS-Fusion \cite{yang2025opengs} & 46.79  & 35.20  & 48.56    & 55.54    & 33.53   & 42.96     \\
DynamicGSG \cite{ge2025dynamicgsg}    & \cellcolor{secondcolor}53.65   & 31.92   & 47.13  & \cellcolor{thirdcolor}57.83   & \cellcolor{secondcolor}42.23   & \textbf{\cellcolor{firstcolor}57.61}   \\
OmniMap \cite{deng2025omnimap}       & \cellcolor{thirdcolor}54.44   & \cellcolor{secondcolor}39.93     & \cellcolor{secondcolor}66.84   & \cellcolor{secondcolor}58.16   & \cellcolor{thirdcolor}42.08     & \cellcolor{thirdcolor}51.84   \\

\textbf{Ours w/o mask IoU} & 51.43    & 
\cellcolor{thirdcolor}36.26    & 
\cellcolor{thirdcolor}54.51   &
53.70    & 
41.58   & 
50.17  \\

\textbf{Ours} & \textbf{\cellcolor{firstcolor}55.48}    & \textbf{\cellcolor{firstcolor}40.53}    & \textbf{\cellcolor{firstcolor}67.94}   & \textbf{\cellcolor{firstcolor}62.26}    & \textbf{\cellcolor{firstcolor}45.51}   & \cellcolor{secondcolor}56.86  \\
 \bottomrule
\end{tabular}
    \caption{Quantitative results of 3D open-vocabulary semantic segmentation on Replica and ScanNet.}
    \label{tab:semseg}
    \vspace{-1em}
\end{table}

\begin{figure}[t]
  \centering
  \includegraphics[width=1.0\linewidth]{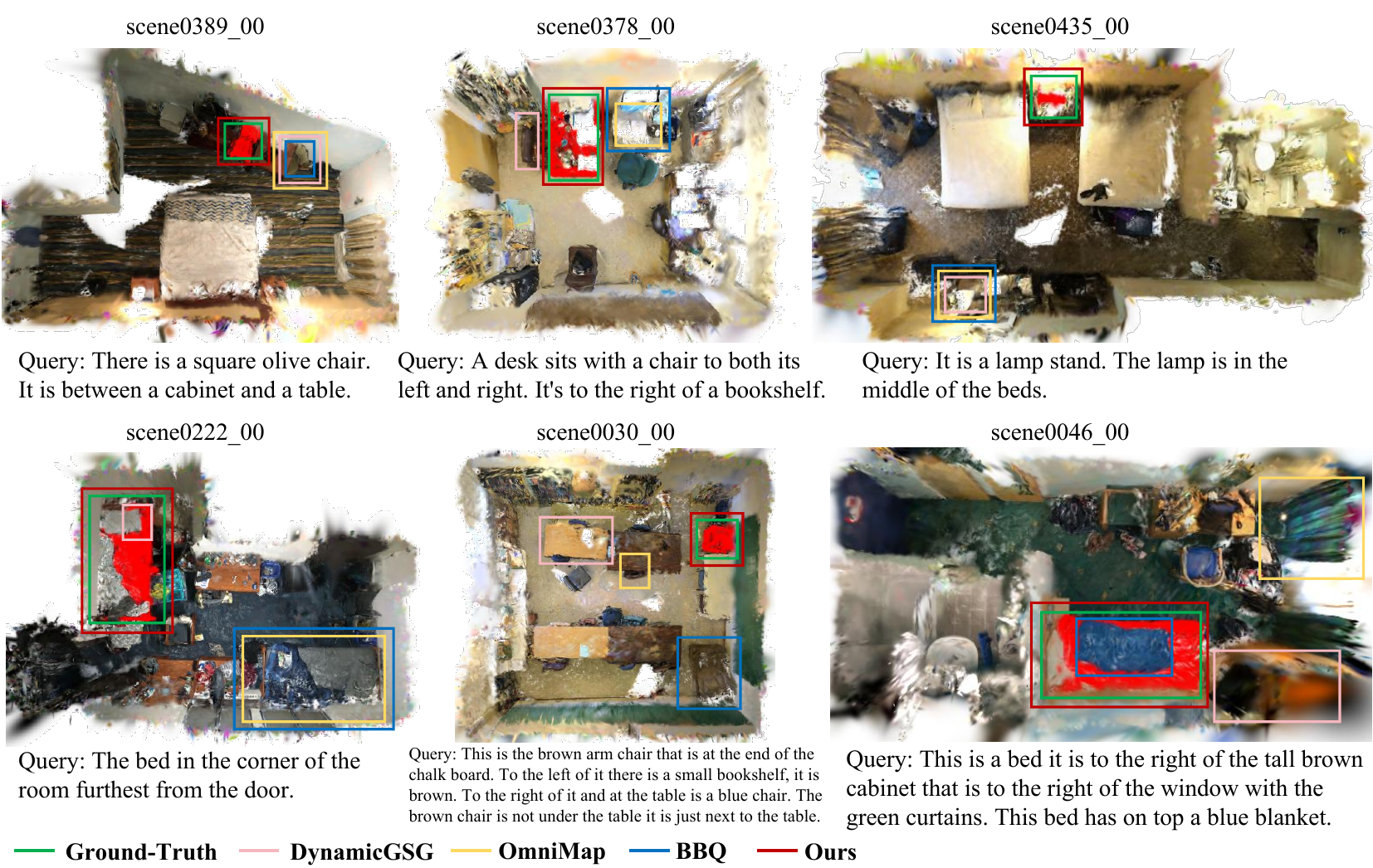}
  \caption{Qualitative results of 3DVG on the ScanRefer and Nr3D.}
  \label{fig:qualitative_3dvg}
  \vspace{-1.5em}
\end{figure}

\begin{table*}[t]
\centering
\scriptsize
\setlength{\tabcolsep}{2.8pt}
\renewcommand{\arraystretch}{1.12}
\resizebox{\textwidth}{!}{%
\begin{tabular}{l c c c c c c c c c c c c c c c c c}
\toprule

\multirow{4.2}{*}{\textbf{Methods}} 
& \multirow{4.2}{*}{\textbf{Agent}} 
& \multicolumn{6}{c}{\textbf{ScanRefer}} 
& \multicolumn{10}{c}{\textbf{Nr3D}} \\

\cmidrule(lr){3-8} \cmidrule(l){9-18}

& 
& \multicolumn{2}{c}{\textbf{Unique (173)}} 
& \multicolumn{2}{c}{\textbf{Multiple (547)}} 
& \multicolumn{2}{c}{\textbf{Overall (720)}} 
& \multicolumn{2}{c}{\textbf{Easy (493)}} 
& \multicolumn{2}{c}{\textbf{Hard (187)}} 
& \multicolumn{2}{c}{\textbf{View-Dep. (166)}} 
& \multicolumn{2}{c}{\textbf{View-Indep. (514)}} 
& \multicolumn{2}{c}{\textbf{Overall (680)}} \\

\cmidrule(lr){3-4} \cmidrule(lr){5-6} \cmidrule(lr){7-8}
\cmidrule(lr){9-10} \cmidrule(lr){11-12} \cmidrule(lr){13-14} \cmidrule(lr){15-16} \cmidrule(l){17-18}

& 
& AP$_{0.25}$ & AP$_{0.5}$ 
& AP$_{0.25}$ & AP$_{0.5}$ 
& AP$_{0.25}$ & AP$_{0.5}$
& AP$_{0.1}$ & AP$_{0.25}$ 
& AP$_{0.1}$ & AP$_{0.25}$ 
& AP$_{0.1}$ & AP$_{0.25}$ 
& AP$_{0.1}$ & AP$_{0.25}$ 
& AP$_{0.1}$ & AP$_{0.25}$ \\

\midrule

\multicolumn{18}{l}{\textbf{Detections:} Zero-Shot \quad / \quad \textbf{PC or Gaussians:} Self-Reconstructed} \\
\cdashmidrule{1-18}

ConceptGraphs \cite{gu2023conceptgraphsopenvocabulary3dscene}
& CLIP 
& 16.18 & 12.14 & 17.18 & 7.50 & 16.94 & 8.61
& 20.28 & 9.74 & 10.70 & 3.21 & 21.69 & 6.02 & 16.34 & 8.56 & 17.65 & 7.94 \\

DynamicGSG \cite{ge2025dynamicgsg}
& CLIP 
& 30.06 & 5.20 & 21.76 & 11.70 & 23.75 & 10.14
& 31.24 & 19.68 & 16.58 & 10.70 & 24.10 & 15.06 & 28.21 & 17.90 & 27.21 & 17.21 \\

OmniMap \cite{deng2025omnimap}
& SBERT 
& \cellcolor{thirdcolor}49.71 & 11.56 & 21.57 & 9.14 & 28.33 & 9.72
& 31.03 & 16.84 & 12.30 & 6.95 & 19.28 & 9.64 & 28.02 & 15.56 & 25.88 & 14.12 \\

BBQ \cite{linok2024barequeriesopenvocabularyobject}
& Qwen2.5-VL
& 29.50 & \textbf{\cellcolor{secondcolor}19.70} & 17.70 & 10.20 & 19.40 & 11.70
& 30.80 & 20.10 & 23.50 & 14.80 & 21.30 & 13.40 & 31.00 & 20.20 & 29.90 & 20.50 \\

\textbf{Ours (text-only)}
& Qwen2.5-VL 
& 34.68 & 16.76 & 26.14 & 13.89 & 28.19 & 14.58 
& 31.24 & 25.15 & 22.99 & 14.44 & 27.71 & 18.67 & 29.38 & 23.35 & 28.97 & 22.21 \\

\textbf{Ours (Target best-views)}
& Qwen2.5-VL
& 35.26 & 17.34 & \cellcolor{thirdcolor}37.84 & \cellcolor{thirdcolor}17.37 & \cellcolor{thirdcolor}37.22 & \cellcolor{thirdcolor}17.36
& \cellcolor{thirdcolor}38.34 & \cellcolor{thirdcolor}27.99 & \cellcolor{thirdcolor}28.88 & \cellcolor{thirdcolor}22.46 & \cellcolor{thirdcolor}34.94 & \cellcolor{thirdcolor}24.10 & \cellcolor{thirdcolor}35.99 & \cellcolor{thirdcolor}27.24 & \cellcolor{thirdcolor}35.74 & \cellcolor{thirdcolor}26.47 \\

\textbf{Ours}
& Qwen2.5-VL 
& \textbf{\cellcolor{secondcolor}53.18} & \cellcolor{thirdcolor}18.50 & \textbf{\cellcolor{secondcolor}39.31} & \textbf{\cellcolor{secondcolor}24.68} & \textbf{\cellcolor{secondcolor}42.64} & \textbf{\cellcolor{secondcolor}23.19}
& \textbf{\cellcolor{secondcolor}48.07} & \textbf{\cellcolor{secondcolor}36.31} & \textbf{\cellcolor{firstcolor}32.09} & \textbf{\cellcolor{secondcolor}24.06} & \textbf{\cellcolor{firstcolor}38.55} & \textbf{\cellcolor{secondcolor}28.92} & \textbf{\cellcolor{secondcolor}45.33} & \textbf{\cellcolor{secondcolor}34.24} & \textbf{\cellcolor{secondcolor}43.68} & \textbf{\cellcolor{secondcolor}32.94} \\

\midrule

\multicolumn{18}{l}{\textbf{Detections:} Preprocessed Mask3D \quad / \quad \textbf{PC:} ScanNet GT} \\
\cdashmidrule{1-18}

SeeGround \cite{li2025seeground}
& Qwen2.5-VL 
& \cellcolor{firstcolor}75.72 & \cellcolor{firstcolor}66.47 & \cellcolor{firstcolor}42.60 & \cellcolor{firstcolor}37.84 & \cellcolor{firstcolor}50.56 & \cellcolor{firstcolor}44.72
& \cellcolor{firstcolor}51.93 & \cellcolor{firstcolor}48.88 & \cellcolor{secondcolor}30.48 & \cellcolor{firstcolor}24.60 & \cellcolor{firstcolor}38.55 & \cellcolor{firstcolor}33.73 & \cellcolor{firstcolor}48.44 & \cellcolor{firstcolor}44.94 & \cellcolor{firstcolor}46.03 & \cellcolor{firstcolor}42.21 \\

\bottomrule
\end{tabular}%
}
\caption{Quantitative results of 3D visual grounding on the ScanRefer~\cite{chen2020scanrefer} and Nr3D~\cite{achlioptas2020referit3d} selected sets. The first group uses zero-shot detections and self-reconstructed point clouds or Gaussian maps, while SeeGround~\cite{li2025seeground} uses preprocessed Mask3D detections and ScanNet ground-truth point clouds as an upper-bound reference. For this controlled visual-input ablation, all three variants use the same query, hierarchical scene-graph serialization, Target/Anchor candidate sets, grounding prompt, and VLM, and differ only in the supplied image input. Ours (text-only) receives no image; Target best-views provide each Target candidate's captured best-view image with the Target ID and currently visible Anchor IDs overlaid; and the full method provides one annotated relation-aware Gaussian view per Target candidate.}
\label{tab:joint_grounding}
\vspace{-1.5em}
\end{table*}


\renewcommand{\arraystretch}{1.2}
\setlength{\tabcolsep}{2.4pt}
\begin{table}[!t]
    \centering
    \begin{tabular}{lcccccc}\toprule
\multirow{2.5}{*}{\textbf{Methods}}      & \multicolumn{3}{c}{\textbf{Replica}} & \multicolumn{3}{c}{\textbf{ScanNet}} \\
\cmidrule(lr){2-4} \cmidrule(lr){5-7} 
& PSNR$\uparrow$        & SSIM$\uparrow$   & LPIPS$\downarrow$  & PSNR$\uparrow$        & SSIM$\uparrow$   & LPIPS$\downarrow$  \\ \midrule 
GS-ICP-SLAM \cite{ha2024rgbdicp}   & 36.92  & 0.968   & 0.042  & 24.11 & 0.803 & 0.363 \\
DynamicGSG \cite{ge2025dynamicgsg}    & 35.62  & \cellcolor{thirdcolor}0.979  & 0.068  & 23.56  & \cellcolor{thirdcolor}0.815  & \cellcolor{secondcolor}0.285  \\
OpenGS-Fusion \cite{yang2025opengs} & \cellcolor{thirdcolor}39.28 & 0.976 & \cellcolor{thirdcolor}0.039 & \cellcolor{secondcolor}26.46 & \cellcolor{secondcolor}0.822 & \textbf{\cellcolor{firstcolor}0.278} \\
OmniMap \cite{deng2025omnimap}      & \cellcolor{secondcolor}39.47  & \cellcolor{secondcolor}0.980  & \cellcolor{secondcolor}0.021 & \cellcolor{thirdcolor}26.14  & 0.814 & 0.295 \\
\textbf{Ours} & \textbf{\cellcolor{firstcolor}40.31}  & \textbf{\cellcolor{firstcolor}0.984}  & \textbf{\cellcolor{firstcolor}0.017} & \textbf{\cellcolor{firstcolor}26.78}  & \textbf{\cellcolor{firstcolor}0.829} & \cellcolor{thirdcolor}0.293 \\
 \bottomrule
\end{tabular}
    \caption{Quantitative comparison of photometric reconstruction quality on the Replica \cite{replica19arxiv} and ScanNet \cite{dai2017scannet} datasets.}
    \label{tab:reconstruction}
    \vspace{-2em}
\end{table}

We evaluate the open-vocabulary semantic segmentation performance of \textit{DGSG-Mind} on Replica~\cite{replica19arxiv} and ScanNet~\cite{dai2017scannet}. For each foreground 3D primitive (point or Gaussian), we compare its semantic feature with the text CLIP\cite{radford2021learningtransferablevisualmodels} features of all object classes in the scene and assign the class with the highest cosine similarity. We report mean Accuracy (mAcc), mean intersection over union (mIoU), and frequency-weighted mIoU (F-mIoU) against ground truth. The instance-association threshold is set to $\tau=0.4$ for Replica and $\tau=0.3$ for ScanNet. As shown in Table~\ref{tab:semseg}, our method achieves the best mAcc and mIoU on both datasets while remaining competitive in F-mIoU on ScanNet. The ablation without mask IoU confirms that rendered mask IoU improves instance fusion and cross-view semantic consistency. In contrast, methods based only on 3D overlap (e.g., ConceptGraphs~\cite{gu2023conceptgraphsopenvocabulary3dscene}) tend to merge small objects into nearby larger ones, methods relying mainly on 2D mask IoU (e.g., DynamicGSG~\cite{ge2025dynamicgsg}) are more sensitive to detector instability, and methods based mainly on voxel statistics (e.g., OpenGS-Fusion~\cite{yang2025opengs}, OmniMap~\cite{deng2025omnimap}) lack sufficient cross-view geometric constraints.

\subsection{3D Visual Grounding}
\label{subsec:object_retrieval}

We evaluate 3D visual grounding on ScanRefer~\cite{chen2020scanrefer} and Nr3D~\cite{achlioptas2020referit3d} queries from the eight selected ScanNet scenes. Semantic-matching baselines use their original vision-language encoders, whereas all agent-based methods use \texttt{Qwen2.5-VL-72B-Instruct}~\cite{bai2025qwen25vltechnicalreport}. Most compared methods, including ours, operate in a zero-shot setting on self-reconstructed geometries. In contrast, SeeGround~\cite{li2025seeground} uses ScanNet ground-truth point clouds and preprocessed Mask3D masks, and is therefore included as an upper-bound reference rather than a directly comparable zero-shot baseline. As shown in Table~\ref{tab:joint_grounding}, our method outperforms the comparable zero-shot baselines on most metrics and substantially surpasses BBQ~\cite{linok2024barequeriesopenvocabularyobject}, the strongest self-reconstructed VLM baseline, which reasons only over the textual scene-graph structure. The controlled ablation shows consistent improvements across both datasets when progressing from text-only reasoning to captured Target best-views and further to relation-aware Gaussian views. These results show that \textit{DGSG-Mind} achieves stronger visual-spatial reasoning by combining ID-annotated Gaussian renderings with structured scene graph context.

\subsection{Photometric Reconstruction Quality}
\label{subsec:reconstruction_quality}

We evaluate the photometric reconstruction on Replica \cite{replica19arxiv} and ScanNet~\cite{dai2017scannet} using standard metrics: PSNR, SSIM, and LPIPS. For fair comparison, all methods use the dataset-provided ground-truth camera poses. As shown in Table~\ref{tab:reconstruction}, \textit{DGSG-Mind} achieves the best results on all Replica metrics and the highest PSNR and SSIM on ScanNet, while remaining competitive in LPIPS. These results demonstrate the effectiveness of voxel-guided Gaussian densification and joint optimization for accurate, photorealistic reconstruction.

\subsection{Real-World Deployment}
\label{subsec:real_world}

\begin{figure}[tb]
  \centering
  \includegraphics[width=1.0\linewidth]{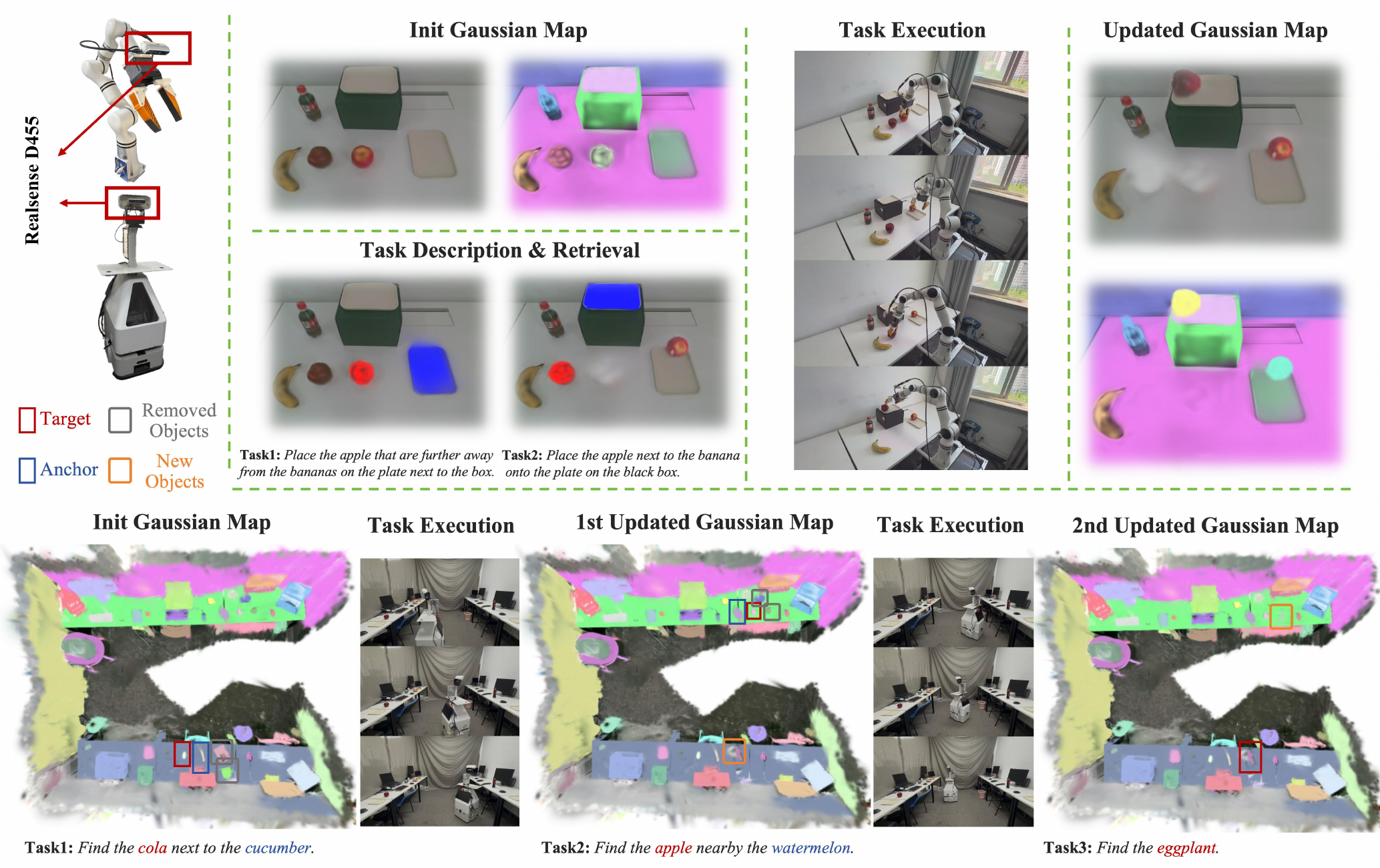}
  \caption{Real-world scene-aware manipulation (top) and target-guided navigation (bottom). \textit{DGSG-Mind} grounds two sequential pick-and-place instructions for arm execution and three navigation targets for the mobile robot as the Gaussian map is updated.}
  \label{fig:real}
  \vspace{-0.5em}
\end{figure}

\begin{table}[!t]
    \centering
    \begingroup
    \footnotesize
    \setlength{\tabcolsep}{3pt}
    \renewcommand{\arraystretch}{1.05}
    \begin{tabular*}{\columnwidth}{@{\extracolsep{\fill}}lcccc@{}}
        \toprule
        \textbf{Variant} & \textbf{Rem.} & \textbf{Mov.} & \textbf{Add.} & \textbf{All (\%)} \\
        \midrule
        DyGSG       & 4/20  & 0/10 & 7/20  & 22.0 \\
        DyGSG+ACE   & 14/20 & 4/10 & 16/20 & 68.0 \\
        \midrule
        Ours (A)       & 15/20 & 4/10 & 17/20 & 72.0 \\
        Ours (A+G)     & 16/20 & 5/10 & 17/20 & 76.0 \\
        Ours (A+S)     & 18/20 & 6/10 & 17/20 & 82.0 \\
        \textbf{Ours (A+G+S)} & \textbf{19/20} & \textbf{7/10} & \textbf{17/20} & \textbf{86.0} \\
        \bottomrule
    \end{tabular*}
    \endgroup
    \caption{\textbf{Dynamic-update success and joint-consistency-term ablation.} Following DynamicGSG's three object-change categories~\cite{ge2025dynamicgsg}, entries report successes/trials for 20 removals, 10 movements, and 20 additions; A, G, and S denote appearance, geometric, and semantic consistency, with A used in all variants. A movement requires successful removal and re-addition without identity preservation. DyGSG uses DynamicGSG's original VIO-based pose initialization, whereas DyGSG+ACE replaces it with ACE.}
    \label{tab:changes}
    \vspace{-2em}
\end{table}

As shown in Fig.~\ref{fig:real}, we deploy \textit{DGSG-Mind} on a tabletop robotic arm and a wheeled mobile robot in real indoor manipulation and navigation scenarios, with RGB-D observations collected by head- and wrist-mounted RealSense D455 cameras. NVIDIA cuVSLAM~\cite{korovko2025cuvslam} provides camera poses for initial map construction, and the posed RGB images are used to train a scene-specific ACE \cite{brachmann2023accelerated} model for subsequent Gaussian-based relocalization. The reconstructed Gaussian maps provide high-quality geometry and semantically consistent instance representations, supporting tabletop manipulation, target-oriented navigation, and long-term scene updates. As shown in Table~\ref{tab:changes}, \textit{DGSG-Mind} substantially outperforms DynamicGSG~\cite{ge2025dynamicgsg} under the same robot-mounted conditions. DynamicGSG relies on continuous VIO and is sensitive to tracking drift during mobile robot motion and close-range arm manipulation, whereas our method independently relocalizes each observation against the optimized Gaussian map. The controlled ablation results further show that ACE substantially strengthens the DynamicGSG baseline, while geometric, appearance, and semantic consistency provide complementary benefits for object-level change detection. By updating only the affected Gaussian instances and scene graph nodes while keeping the static background fixed, \textit{DGSG-Mind} maintains an up-to-date scene representation for long-term embodied tasks.

\section{Conclusion}
\label{sec:conclusion}

\textit{DGSG-Mind} advances long-term embodied scene understanding in dynamic environments through hybrid instance-aware reconstruction, localized dynamic updates, and multimodal reasoning. Extensive experiments on public benchmarks and real-world platforms validate its effectiveness for scene understanding and task execution. Nevertheless, the system still relies on SLAM pose accuracy for initial reconstruction and ACE training, and its scalability to large-scale outdoor scenes is limited by the storage and GPU memory costs of 3D Gaussians. Future work will focus on developing an integrated tracking module and more scalable Gaussian map representations.

\bibliographystyle{IEEEtran}
\bibliography{references}

\clearpage
\setcounter{section}{0}
\setcounter{subsection}{0}
\setcounter{figure}{0}
\setcounter{table}{0}
\setcounter{algorithm}{0}
\setcounter{equation}{0}
\renewcommand{\thefigure}{S\arabic{figure}}
\renewcommand{\thetable}{S\arabic{table}}
\renewcommand{\thealgorithm}{S\arabic{algorithm}}
\renewcommand{\theequation}{S\arabic{equation}}
\makeatletter
\def\theHsection{supp.\arabic{section}}
\def\theHsubsection{supp.\arabic{section}.\arabic{subsection}}
\def\theHfigure{supp.\arabic{figure}}
\def\theHtable{supp.\arabic{table}}
\def\theHalgorithm{supp.\arabic{algorithm}}
\def\theHequation{supp.\arabic{equation}}
\begingroup
\let\@maketitle\@oldmaketitle
\def\@title{\LARGE\bfseries DGSG-Mind: Supplementary Material}
\def\@author{}
\def\@thanks{}
\ArxivSupplementMaketitle
\endgroup
\makeatother
\thispagestyle{empty}
\pagestyle{empty}
\raggedbottom

\section{Complete Implementation Configuration}
\label{sec:supp-implementation}
The implementation settings are summarized as follows: 

\begin{itemize}
    \item \textbf{Voxel grid.} The voxel resolution is \(0.03\,\mathrm{m}\).

    \item \textbf{Cross-modal instance association.} The geometric, rendered-mask IoU, and semantic terms use weights \(\lambda_1=0.4\), \(\lambda_2=0.4\), and \(\lambda_3=0.2\), respectively. The association threshold is set to \(\tau=0.4\) for Replica and \(\tau=0.3\) for ScanNet and real-world experiments.

    \item \textbf{Dynamic change detection.} An instance is evaluated only when more than \(50\%\) of its associated Gaussians are visible in the current frustum. The valid-depth threshold is \(\tau_d=0.05\,\mathrm{m}\); the geometric, appearance, and semantic consistency weights are \(\lambda_{10}=0.2\), \(\lambda_{11}=0.4\), and \(\lambda_{12}=0.4\); and the final dynamic change threshold is \(\delta_{\mathrm{change}}=0.35\).

    \item \textbf{Keyframe selection.} A keyframe is selected every \(\delta_n=5\) frames or when the camera translation exceeds \(0.1\,\mathrm{m}\) or its rotation exceeds \(0.1\,\mathrm{rad}\). For joint multi-view optimization, \(\delta_m=10\) keyframes are sampled.

    \item \textbf{Optimization iterations.} For Gaussian mapping, each keyframe update is optimized for \(10\) iterations, with \(10\) historical keyframes sampled for each multi-view optimization. Gaussian-based camera pose refinement and localized dynamic refinement each use \(200\) iterations.

    \item \textbf{VLM deployment.} Because the Qwen2.5-VL API service was unavailable during the revision, the open-source Qwen2.5-VL-72B-Instruct model was self-hosted on a cloud server with eight NVIDIA A100 GPUs (80~GB each). All new ablations and timing measurements involving Qwen inference use this deployment; decoding follows the model's default generation configuration.
\end{itemize}

\section{Relation-aware RoI View Generation}
\label{sec:supp-roi}

\begin{algorithm}[!t]
\caption{$\operatorname{LookAtFit}$: relation-aware RoI view generation}
\label{alg:supp-lookatfit}
\begin{algorithmic}[1]
\Require Target candidates $\mathcal{C}_T$; Anchor candidate sets $\{\mathcal{C}_A(q)\}_{q\in\mathcal{Q}_{\mathrm{anchor}}}$; node centers $\mathbf{p}_j$ and 3D bboxes $\mathbf{b}_j$; Gaussian map $\mathcal{G}$; Target best-view poses $\nu_T=[\mathbf{R}_{\mathrm{ref}}\mid\mathbf{t}_{\mathrm{ref}}]$; intrinsics $\mathbf{K}$; image size $(W,H)$; scene XY bounds $\Omega_{xy}$; scene-up vector $\widehat{\mathbf{u}}_s$
\Ensure Annotated relation-aware views $\{\mathbf{I}_T\}_{o_T\in\mathcal{C}_T}$
\ForAll{$o_T\in\mathcal{C}_T$}
    \State $\mathcal{A}_T\gets\operatorname{Unique}\!\left(\bigcup_{q\in\mathcal{Q}_{\mathrm{anchor}}}\mathcal{C}_A(q)\right)\setminus\{o_T\}$
    \State $\mathcal{R}_T\gets\{o_T\}\cup\mathcal{A}_T$
    \If{$|\mathcal{A}_T|>0$} 
        \State $\mathbf{q}_T\gets0.7\,\mathbf{p}_T+\dfrac{0.3}{|\mathcal{A}_T|}\sum_{a\in\mathcal{A}_T}\mathbf{p}_a$
    \Else
        \State $\mathbf{q}_T\gets\mathbf{p}_T$
    \EndIf
    \State $\mathcal{V}_T\gets\bigcup_{o\in\mathcal{R}_T}\operatorname{BBoxVertices}(o)$
    \State $\boldsymbol{\mu}_T\gets|\mathcal{V}_T|^{-1}\sum_{\mathbf{v}\in\mathcal{V}_T}\mathbf{v}$; \quad $\rho_T\gets\max_{\mathbf{v}\in\mathcal{V}_T}\|\mathbf{v}-\boldsymbol{\mu}_T\|_2$
    \State $\mathbf{p}_{\mathrm{ref}}\gets-\mathbf{R}_{\mathrm{ref}}^{\top}\mathbf{t}_{\mathrm{ref}}$
    \State $\mathbf{p}^{(0)}\gets\mathbf{p}_{\mathrm{ref}}+\max(0.2,0.35\rho_T)\widehat{\mathbf{u}}_s$
    \State $k^{\star}\gets\varnothing$
    \For{$k\gets0$ \textbf{to} $7$}
        \State $\mathbf{T}^{(k)}\gets\operatorname{LookAt}(\mathbf{q}_T,\mathbf{p}^{(k)},\widehat{\mathbf{u}}_s)$
        \If{$\operatorname{FitInImage}(\mathcal{V}_T,\mathbf{T}^{(k)},\mathbf{K},W,H,0.08)$}
            \State $k^{\star}\gets k$; \textbf{break}
        \EndIf
        \State $\widetilde{\mathbf{p}}^{(k+1)}\gets\mathbf{q}_T+1.15(\mathbf{p}^{(k)}-\mathbf{q}_T)$
        \If{$\widetilde{\mathbf{p}}_{xy}^{(k+1)}\in\Omega_{xy}$}
            \State $\mathbf{p}^{(k+1)}\gets\widetilde{\mathbf{p}}^{(k+1)}$
        \Else
            \State $k^{\star}\gets k$; \textbf{break}
        \EndIf
    \EndFor
    \If{$k^{\star}=\varnothing$}
        \State $k^{\star}\gets8$; \quad $\mathbf{T}^{(8)}\gets\operatorname{LookAt}(\mathbf{q}_T,\mathbf{p}^{(8)},\widehat{\mathbf{u}}_s)$
    \EndIf
    \State $\mathbf{I}_T\gets\operatorname{Render}(\mathcal{G},\mathbf{K},\mathbf{T}^{(k^{\star})})$
    \State $\mathbf{I}_T\gets\operatorname{Annotate}(\mathbf{I}_T,o_T\!:\!\mathrm{red},\mathcal{A}_T\!:\!\mathrm{blue})$
\EndFor
\end{algorithmic}
\end{algorithm}

Given a parsed phrase $q$, the candidate score is $s(q,o_j)=\hat{\mathbf e}_q^{\top}\hat{\mathbf F}_j$, between the normalized CLIP feature of the phrase and the normalized semantic feature of the map instance associated with node $o_j$. Category and caption metadata are reserved for final grounding. Target and Anchor phrases are retrieved independently, with up to five nodes satisfying $s(q,o_j)\geq\max(0.2,s_{\max}(q)-0.1)$. Scene-graph relations help the parser identify the phrases and are preserved for final reasoning, but do not enter this cosine score or candidate filtering. Algorithm~\ref{alg:supp-lookatfit} generates exactly one relation-aware view per Target candidate. For a Target $o_T$, all distinct retrieved Anchors with different instance IDs are merged into the same joint region; thus $N_T$ Target candidates produce $N_T$ views, rather than one view for every Target--Anchor pair. The stored Target best-view pose is only the scene-consistent initialization: the final virtual camera may translate backward and re-orient to improve joint Target--Anchor coverage, so no additional capture-distance requirement is imposed. Here, $\operatorname{FitInImage}$ checks positive depth and projection inside an 8\% image margin, so the procedure seeks joint coverage but does not guarantee complete containment for every scene configuration. $\operatorname{LookAt}(\mathbf{q},\mathbf{p},\widehat{\mathbf{u}})$ returns the world-to-camera extrinsic obtained by calling Open3D's \texttt{scene.camera.look\_at(center=q, eye=p, up=u)} with look-at point $\mathbf{q}$, camera center $\mathbf{p}$, and scene-up direction $\widehat{\mathbf{u}}$. The rendered image is annotated with compact red Target and blue Anchor IDs. These IDs act as correspondence pointers to the candidate-node records; the VLM additionally receives the original query, serialized scene graph, candidate sets, and fine-grained node captions, so reasoning is not based on labels and spatial positions alone.

\newcommand{\suppprompttemplates}{%
\section{Prompt Templates}
\label{sec:supp-prompts}

The three fixed templates correspond to (i) category and functional-role assignment from a masked best-view caption, (ii) Target--Anchor parsing from a referring expression, and (iii) final grounding from the query, annotated relation-aware views, candidate sets, and serialized scene graph. The first two stages return strict JSON; the final stage selects an ID from the Target set and explains the decision. Fine-grained captions preserve color, material, shape, and distinctive components when an ID label covers part of a small object. Runtime values are inserted only into fields enclosed in angle brackets. Decoding follows the model's default generation configuration without overriding temperature, top-$p$, or other sampling parameters.

\begin{tcolorbox}[colback=black!3, colframe=black!35, boxrule=0.6pt, top=2pt, bottom=2pt]
\footnotesize\raggedright
\textbf{Prompt 2: Target--Anchor query parsing}

\smallskip
\textbf{Task.} You are an assistant designed to identify relationships between objects in a scene. Determine the \emph{Target}, the object being located, and the \emph{Anchor}(s), the reference object(s) providing spatial or contextual information.

\smallskip
\textbf{Representative examples.}

\textbf{Query:} \texttt{Find the chair that is next to the wooden table.}

\textbf{Output:} \texttt{\{``Target'': ``chair'', ``Anchor'': ``wooden table''\}}

\textbf{Query:} \texttt{The largest table in the room.}

\textbf{Output:} \texttt{\{``Target'': ``table'', ``Anchor'': ``''\}}

\textbf{Query:} \texttt{The office chair that is between the desk and the sofa.}

\textbf{Output:} \texttt{\{``Target'': ``office chair'', ``Anchor'': [``desk'', ``sofa'']\}}

\smallskip
\textbf{Output rule.} Return exactly one JSON object. Use a string for one Anchor, a list for multiple Anchors, and an empty string when no Anchor is present.

\smallskip
\textbf{Query:} \texttt{<free-form referring expression>}
\end{tcolorbox}

\begin{tcolorbox}[colback=black!3, colframe=black!35, boxrule=0.6pt, top=2pt, bottom=2pt]
\footnotesize\raggedright
\textbf{Prompt 3: Final grounding}

\smallskip
\textbf{Task.} Identify the object that best satisfies the referring expression from the serialized 3D scene information and, when available, the annotated image inputs.

\smallskip
\textbf{Inputs.}

\texttt{<annotated image input(s), if provided>}\\
\texttt{<hierarchical scene graph: instance ID, category, description, role, dimensions, 3D center, and children>}\\
\textbf{Target set:} \texttt{<retrieved Target candidates>}\\
\textbf{Anchor set:} \texttt{<retrieved Anchor candidates>}\\
\textbf{Query:} \texttt{<free-form referring expression>}

\smallskip
\textbf{Reasoning rules.} Treat the $X$- and $Y$-axes as horizontal and the $Z$-axis as vertical. Evaluate appearance attributes and view-dependent relations from the images when available, and use the 3D coordinates for metric spatial relations. Select the final ID only from the Target set. For each Target candidate, use retrieved Anchor candidates other than the same instance as relational context. Do not guess when the evidence is insufficient.

\smallskip
\textbf{Response format:}\\
\texttt{Predicted ID: <ID>}\\
\texttt{Explanation: <why this target is selected and other candidates are rejected>}
\end{tcolorbox}

\begin{tcolorbox}[colback=black!3, colframe=black!35, boxrule=0.6pt, top=2pt, bottom=2pt]
\footnotesize\raggedright
\textbf{Prompt 1: Joint object-category and functional-role assignment}

\smallskip
\textbf{Task.} You are an assistant designed to identify the main object category from a detailed object description and assign its functional role in a 3D scene graph.

\smallskip
\textbf{Rules.}
\begin{enumerate}
\item Identify the main object and express its category using a concise noun phrase.
\item Ignore secondary objects mentioned only to describe the main object.
\item Assign exactly one of the following functional roles to the main object:
\begin{itemize}
\item \textbf{Asset:} Large, functional furniture that is typically placed on the floor and can support people or other objects, or provide storage. Examples include chairs, tables, cabinets, sofas, and beds. Do not classify furniture parts, small portable containers, lamps, vases, or decorations as Assets.
\item \textbf{Standalone:} An independently rooted object or fixture that should be connected directly to the scene root rather than organized under an Asset. This includes objects suspended from the ceiling, objects embedded in a wall, and independently placed non-Asset objects such as stools and trash bins.
\item \textbf{Ordinary:} Any remaining object, typically a smaller or movable object that may be placed on or contained in an Asset.
\end{itemize}
\item First determine the main object category, and then assign the functional role to that same object.
\end{enumerate}

\textbf{Output format.} Return exactly one JSON object without additional explanation:

\texttt{\{``Object category'': ``<main object>'', ``Functional role'': ``<Asset|Standalone|Ordinary>''\}}
\end{tcolorbox}
}

\section{Association Sensitivity}
\label{sec:supp-sensitivity}

\begin{table}[t]
\centering
\scriptsize
\setlength{\tabcolsep}{5pt}
\renewcommand{\arraystretch}{1.04}
\begin{tabular}{@{}lrrr@{}}
\toprule
\multicolumn{4}{c}{\textbf{Replica}} \\
\cmidrule(lr){1-4}
\textbf{Setting} & \textbf{mAcc} & \textbf{mIoU} & \textbf{F-mIoU} \\
\midrule
$\boldsymbol{\lambda}=(0.20,0.60,0.20)$ & 51.88 & 36.73 & 64.86 \\
$\boldsymbol{\lambda}=(0.60,0.20,0.20)$ & 52.15 & 37.15 & 66.43 \\
$\boldsymbol{\lambda}=(0.45,0.45,0.10)$ & 53.31 & 38.55 & 66.57 \\
$\boldsymbol{\lambda}=(0.30,0.30,0.40)$ & 52.12 & 36.91 & 64.95 \\
$\tau=\tau_0+0.1$ & 52.43 & 37.99 & 65.22 \\
$\tau=\tau_0-0.1$ & 53.35 & 38.90 & 66.45 \\
\textbf{Default} & \textbf{55.48} & \textbf{40.53} & \textbf{67.94} \\
\midrule
\multicolumn{4}{c}{\textbf{ScanNet}} \\
\cmidrule(lr){1-4}
\textbf{Setting} & \textbf{mAcc} & \textbf{mIoU} & \textbf{F-mIoU} \\
\midrule
$\boldsymbol{\lambda}=(0.20,0.60,0.20)$ & 58.86 & 41.85 & 50.54 \\
$\boldsymbol{\lambda}=(0.60,0.20,0.20)$ & 60.27 & 43.84 & 54.11 \\
$\boldsymbol{\lambda}=(0.45,0.45,0.10)$ & 61.58 & 45.11 & 54.02 \\
$\boldsymbol{\lambda}=(0.30,0.30,0.40)$ & 51.50 & 35.24 & 44.36 \\
$\tau=\tau_0+0.1$ & 60.32 & 44.18 & 53.87 \\
$\tau=\tau_0-0.1$ & 56.92 & 40.71 & 47.81 \\
\textbf{Default} & \textbf{62.26} & \textbf{45.51} & \textbf{56.86} \\
\bottomrule
\end{tabular}
\caption{Association-parameter sensitivity.}
\label{tab:supp-sensitivity}
\vspace{-2em}
\end{table}
The weight order is geometry, rendered-mask IoU, and semantics. Geometry and rendered-mask IoU use equal weights because they provide complementary occupancy and view-consistent silhouette evidence; semantic similarity is retained as a lower-weight auxiliary cue because it is more affected by category ambiguity and domain differences. The default threshold $\tau_0$ is 0.4 for Replica and 0.3 for ScanNet; all other settings are fixed. As shown in Table \ref{tab:supp-sensitivity}, the default is best on all six metrics. With semantic weight fixed at 0.2, shifting weight away from rendered-mask IoU lowers mIoU by 3.38 and 1.67 points on Replica and ScanNet, while shifting weight toward IoU lowers it by 3.80 and 3.66 points. Increasing the semantic weight to 0.4 causes the largest ScanNet degradation, from 45.51 to 35.24 mIoU. Across the tested settings, Replica has smaller ranges of mAcc/mIoU/F-mIoU (3.60/3.80/3.08 points) than ScanNet (10.76/10.27/12.50 points). This higher ScanNet sensitivity reflects noisier depth, masks, and reconstruction; it motivates using $\tau=0.3$ there and in real-world data, while lower thresholds introduce excessive incorrect associations.

\section{Runtime and Resource Usage}
\label{sec:supp-runtime}

\begin{table}[t]
\centering
\scriptsize
\setlength{\tabcolsep}{3pt}
\renewcommand{\arraystretch}{1.04}
\begin{tabular}{@{}lcc@{}}
\toprule
\textbf{Operation} & \textbf{Runs} & \textbf{Time (s)} \\
\midrule
Camera relocalization & 20 & $1.08\pm0.08$ \\
Dynamic update & 20 & $7.69\pm1.61$ \\
3D query (ScanRefer) & 20 & $14.31\pm5.29$ \\
3D query (real world) & 20 & $6.38\pm1.53$ \\
ACE training / scene & 10 & $105.1$ \\
\bottomrule
\end{tabular}

\vspace{0.5em}

\begin{tabular}{@{}lccccc@{}}
\toprule
\textbf{Dataset} & \textbf{$N$} & \textbf{GPU$^\dagger$} & \textbf{s/f} & \textbf{FPS} & \textbf{MB} \\
\midrule
Replica & 8 & $21.92/22.23$ & 0.85 & 1.17 & 12.05 \\
ScanNet & 8 & $19.07/22.26$ & 0.98 & 1.02 & 21.92 \\
RealSense & 10 & $21.74/22.31$ & 1.06 & 0.94 & 12.14 \\
\bottomrule
\end{tabular}
\caption{Runtime and resources. $^\dagger$Mean/maximum peak GPU memory in GiB; MB is persistent map storage per scene. ACE training uses approximately 2.0 GiB.}
\label{tab:supp-runtime-resources}
\vspace{-1.5em}
\end{table}

Latency is reported as mean $\pm$ sample standard deviation over 20 runs. Local DGSG-Mind operations use an NVIDIA RTX 4090D workstation. Qwen inference for Qwen-based stages, including metadata generation and end-to-end 3D queries, uses the self-hosted open-source Qwen2.5-VL-72B-Instruct deployment on a cloud server with eight NVIDIA A100 GPUs (80~GB each); the corresponding measurements include network/service latency. Relocalization covers ACE-based coarse-pose estimation followed by Gaussian pose refinement; dynamic update covers change detection, localized Gaussian refinement, and scene-graph synchronization; query latency covers the complete language-to-grounded-result pipeline.

The reconstruction throughput is approximately one frame per second across synthetic, benchmark, and real RGB-D inputs. The persistent map separates Gaussian geometry, appearance, opacity, scale, rotation, and instance assignments in the 3DGS representation from instance IDs, semantic features, categories, captions, functional roles, and best-view metadata in the instance summary. Persistent storage is dominated by the Gaussian representation and grows with the number of reconstructed Gaussians, whereas GPU memory is the primary current scalability bottleneck because the peak measurement includes loaded models, the scene representation, and intermediate buffers. ACE is trained once per scene from posed RGB images; its scene-specific prediction head provides an independent coarse pose for later revisits, after which the Gaussian map performs local refinement. This avoids propagating accumulated VIO drift during subsequent updates, but does not remove the need for posed data during initial reconstruction and ACE training.

\section{Failure Analysis}
\label{sec:supp-failure}

\begin{figure}[!t]
\centering
\setlength{\tabcolsep}{1pt}
\renewcommand{\arraystretch}{0.85}
\begin{tabular}{ccc}
\multicolumn{3}{c}{\scriptsize\textbf{Observation 1}} \\
\includegraphics[width=0.30\columnwidth]{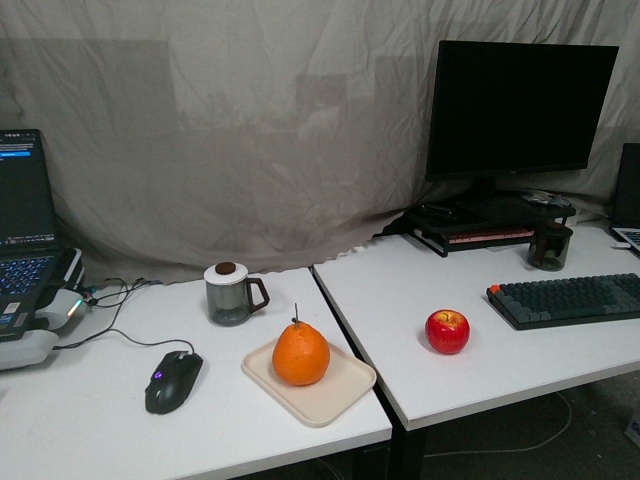} &
\includegraphics[width=0.30\columnwidth]{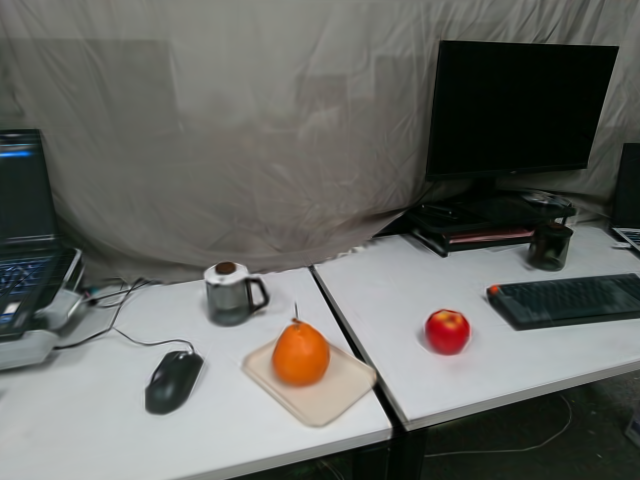} &
\includegraphics[width=0.30\columnwidth]{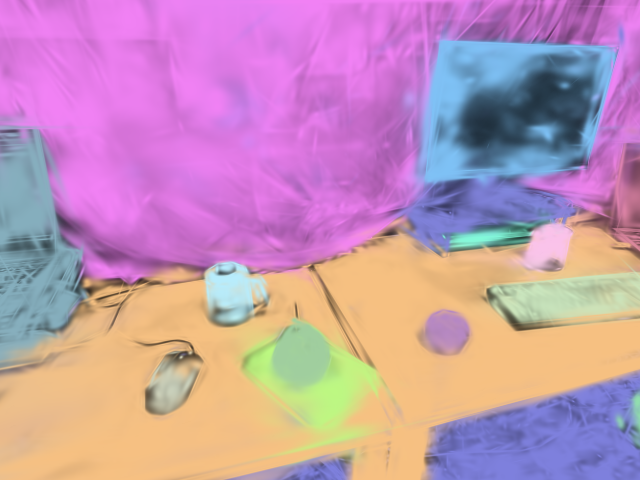} \\
\multicolumn{3}{c}{\scriptsize\textbf{Observation 2}} \\
\includegraphics[width=0.30\columnwidth]{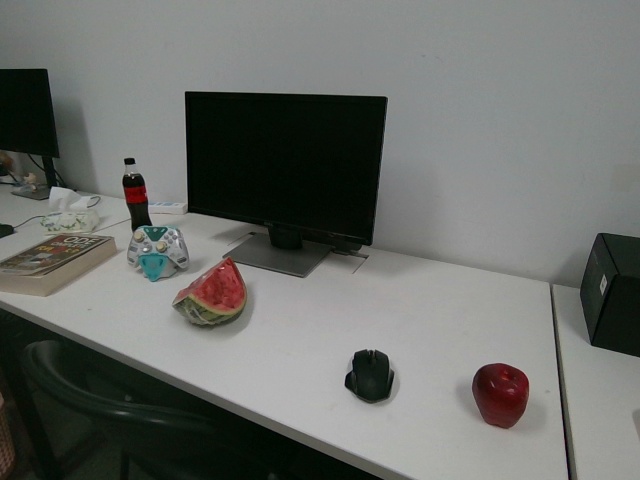} &
\includegraphics[width=0.30\columnwidth]{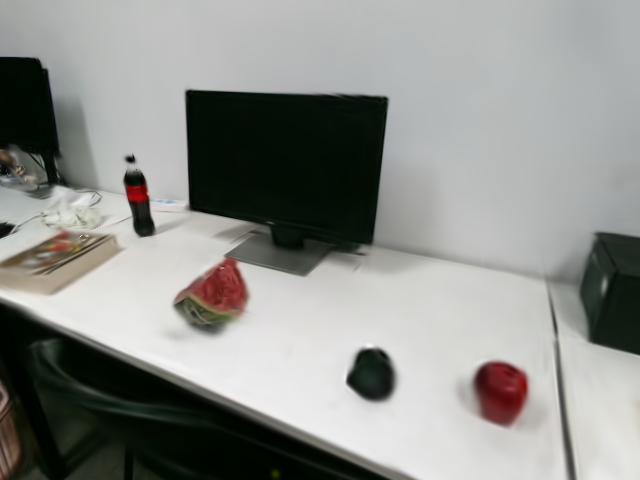} &
\includegraphics[width=0.30\columnwidth]{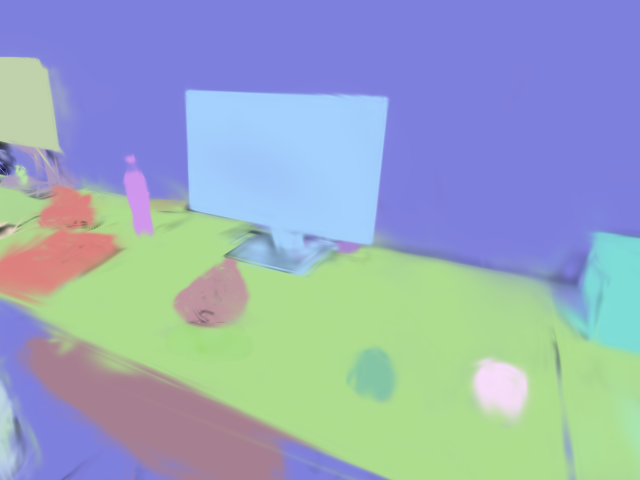} \\
\end{tabular}
\caption{Representative missed-detection failure (columns: RGB observation, updated RGB, updated instance). The controller is removed in Observation 1 but missed by YOLO-World after reappearing in Observation 2.}
\label{fig:supp-failure}
\vspace{-1.5em}
\end{figure}

The full system succeeds in 43 of 50 real-robot object-change trials across 13 scenes. Each scene retains its Gaussian map, scene graph, and scene-specific ACE model through at most four successive consistency-check/update invocations; there is no map reconstruction, graph reset, or ACE retraining between checks. The seven failures comprise one removal, three movements, and three additions. An instance is evaluated only when more than 50\% of its associated Gaussians are visible; this conservative gate defers decisions under partial visibility and reduces false removals, but can also defer updates for large or heavily occluded objects. The representative failure in Fig.~\ref{fig:supp-failure} is caused by a single-view YOLO-World miss: the controller is removed correctly but not re-added after it reappears. The cucumber and watermelon are updated correctly in the same sequence. In the current protocol, movement is handled as removal followed by re-addition with a new identity, so success requires both operations. Multi-view or temporal evidence accumulation could reduce such detector misses.

\suppprompttemplates

\end{document}